\documentclass[10pt,twocolumn,letterpaper]{article}

\usepackage{cvpr}

\usepackage[normalem]{ulem}
\usepackage{soul}

\usepackage{xspace}
\usepackage{lipsum}
\usepackage[dvipsnames]{xcolor}
\usepackage{adjustbox}
\newcommand{\funfact}{\textit{FunFact}\xspace}
\newcommand{\funthor}{\textit{FunThor}\xspace}
\definecolor{ReneBlue}{RGB}{219, 231, 252}
\usepackage{subcaption}

\renewcommand{\paragraph}[1]{\vspace{.0em}\noindent\textbf{#1}}

\usepackage{soul}
\usepackage{booktabs}
\usepackage{multirow}
\usepackage{graphicx}
\usepackage{subcaption}
\usepackage{xcolor}
\usepackage{tikz}
\usepackage[most]{tcolorbox}
\usepackage{algorithm}
\usepackage{algpseudocode}
\usepackage{titletoc}

\DeclareRobustCommand{\legendbox}[1]{
  \tikz[baseline=0.15em] \draw[fill=#1,draw=black] (0,0) rectangle (0.9em,0.9em);
}
\DeclareRobustCommand{\legendcircle}[1]{
  \tikz[baseline=-0.3em] \draw[fill=#1,draw=black] (0,0) circle (0.45em);
}

\definecolor{objectgreen}{HTML}{d5e8d4}
\definecolor{objectpurple}{HTML}{e1d5e7}

\definecolor{cvprblue}{rgb}{0.21,0.49,0.74}
\usepackage[pagebackref,breaklinks,colorlinks,allcolors=cvprblue]{hyperref}

\def\paperID{N/A}
\def\confName{CVPR}
\def\confYear{2026}

\title{FunFact: Building Probabilistic Functional 3D Scene Graphs via \\ Factor-Graph Reasoning}

\author{
Zhengyu Fu$^{1}$ \quad
René Zurbrügg$^{1}$ \quad
Kaixian Qu$^{1}$ \quad
Marc Pollefeys$^{1, 2}$ \\
Marco Hutter$^{1}$ \quad
Hermann Blum$^{3\dagger}$ \quad
Zuria Bauer$^{1\dagger}$ \\
\\
$^{1}$ETH Zürich \quad $^{2}$Microsoft \quad
$^{3}$University of Bonn \& Lamarr Institute\\
{\small $^\dagger$Equal supervision}
\vspace{-5mm}
}

\begin{document}
\maketitle
\begin{abstract}
Recent work in 3D scene understanding is moving beyond purely spatial analysis toward functional scene understanding.
However, existing methods often consider functional relationships between object pairs in isolation, failing to capture the scene-wide interdependence that humans use to resolve ambiguity.
We introduce \funfact, a framework for constructing probabilistic open-vocabulary functional 3D scene graphs from posed RGB-D images. \funfact first builds an object- and part-centric 3D map and uses foundation models to propose semantically plausible functional relations. These candidates are converted into factor graph variables and constrained by both LLM-derived common-sense priors and geometric priors. This formulation enables joint probabilistic inference over all functional edges and their marginals, yielding substantially better calibrated confidence scores. To benchmark this setting, we introduce \funthor, a synthetic dataset based on AI2-THOR with part-level geometry and rule-based functional annotations. Experiments on SceneFun3D, FunGraph3D, and \funthor show that \funfact improves node and relation discovery recall and significantly reduces calibration error for ambiguous relations, highlighting the benefits of holistic probabilistic modeling for functional scene understanding. See our project page at \href{https://funfact-scenegraph.github.io/}{\texttt{https://funfact-scenegraph.github.io/}}.

\end{abstract}
\vspace{-5mm}
\section{Introduction}
\label{sec:intro}

Understanding 3D environments through their \emph{functional} relationships, rather than only geometry or semantics, is increasingly recognized as a key frontier in computer vision~\cite{delitzas2024scenefun3d, rotondi2025fungraph, zhang2025open}. Functional scene understanding seeks to capture how entities interact: which elements control others, which parts afford action, and how one can interact within the scene. Such information is crucial for virtual training systems~\cite{kolve2017ai2, puig2018virtualhome}, AR assistive systems providing actionable guidance~\cite{li2025satori}, and robots that must reason about acting in everyday environments~\cite{pmlr-v205-ichter23a, huang2022language, pmlr-v229-rana23a, ye2017can}, deciding \emph{what} is present and \emph{how} it can be used.

\begin{figure}[t!]
    \centering
    \includegraphics[width=1.0\linewidth]{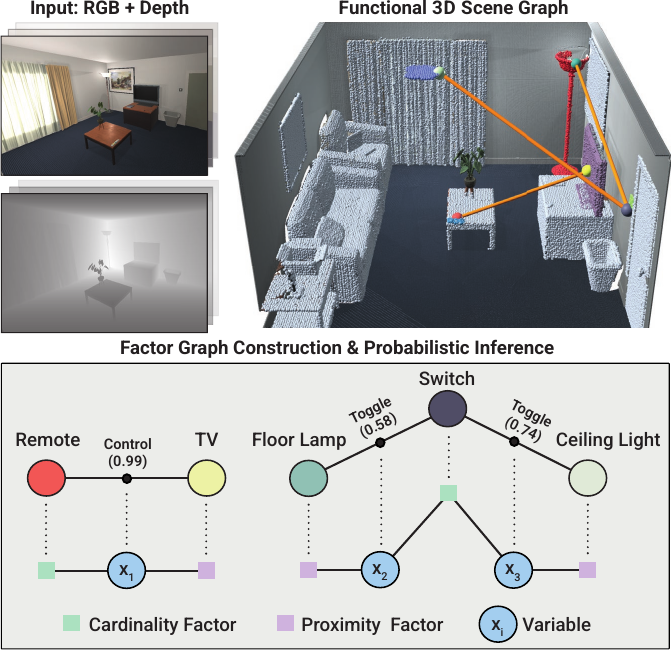}
    \caption{\textbf{\funfact for functional scene understanding.} Given posed RGB-D inputs, \funfact reconstructs an object- and part-centric 3D map and builds a functional scene graph (top). Candidate relations (e.g., \textit{remote controls TV}, \textit{switch toggles lamp}) are encoded as binary variables in a dual factor graph (bottom), where cardinality and proximity factors jointly resolve ambiguities via belief propagation, yielding calibrated per-edge confidence scores.}
    \label{fig:teaser} \vspace{-6mm}
\end{figure}

At its core, functionality can be viewed as a set of relations between entities in the scene. For example, we may want to know which light switch turns on which lamp, which knob controls which burner, or which cable must be unplugged to power down the electric kettle. However, many of these relations are not fully observable from vision alone: the light and its switch may be far apart or occluded, and the causal effect of toggling the switch is not directly encoded in static appearance. This inherent gap between visual evidence and functional behavior makes functional scene understanding particularly challenging. It calls for models that can combine geometric and semantic cues with structured priors and external knowledge to infer plausible functional relationships under ambiguity. Because functionality remains ambiguous from static observations, we argue that prior models should not just predict the most likely connections, but also model the distribution over all likely options.
To this end, we propose \funfact, which predicts calibrated per-functional-edge confidence scores holistically via factor graph optimization. Like prior work~\cite{rotondi2025fungraph, zhang2025open}, we use foundation models as semantic priors to construct open-vocabulary functional 3D scene graphs from posed RGB-D images. By jointly modeling functional relationships and their uncertainties, however, \funfact captures scene interdependencies that independent pairwise reasoning misses.

We validate our method through extensive experiments on both real-world datasets (SceneFun3D~\cite{delitzas2024scenefun3d}, FunGraph3D~\cite{zhang2025open}) and a newly annotated synthetic dataset \funthor collected from AI2-THOR~\cite{kolve2017ai2} scenes. The results demonstrate significant improvements in node and relation discovery recall compared to strong baselines, along with substantially lower expected calibration error (ECE)~\cite{guo2017calibration} in confidence estimation, highlighting \funfact's advantages of holistic probabilistic modeling for functional scene understanding. In summary, our main contributions are:
\begin{itemize}
    \item A novel and robust pipeline for reconstructing open-vocabulary functional 3D scene graphs from posed RGB-D inputs.
    \item A new synthetic dataset (\funthor) for benchmarking functional scene understanding with part-level geometry and dense functional annotations.
    \item A factor-graph formulation that combines LLM priors with geometric evidence to jointly infer functional relations and produce better-calibrated confidence estimates.
\end{itemize}

\section{Related Work}
\label{sec:related_work}

\paragraph{3D Semantic Representations.}
Recent methods leverage foundation models~\cite{caron2021emerging, kirillov2023segment, openai2023gpt4technicalreport} to produce open-vocabulary scene representations without task-specific retraining. OpenScene~\cite{Peng_2023_CVPR} fuses multi-view 2D CLIP-aligned~\cite{radford2021learning} features with distilled 3D point features for zero-shot semantic retrieval, while
ConceptFusion~\cite{conceptfusion} extends this to multimodal queries including text, clicks, images, and audio. Shifting toward object-centric views, OpenMask3D~\cite{takmaz2023openmask3d} aggregates 3D instance mask features across views. Other approaches, such as Tag Map~\cite{pmlr-v270-zhang25e}, build large-vocabulary semantic maps by combining large-scale image tagging~\cite{Zhang_2024_CVPR} with coarse space carving~\cite{kutulakos2000theory}.
Our pipeline builds on these ideas, using similar semantic grounding to identify object and part candidates before reasoning about higher-level functional structure.

\paragraph{3D Scene Graphs.}
3D scene graphs (3DSGs) represent environments as structured graphs of entities and their relationships.
Early works~\cite{armeni20193d, kim20203d} unified geometry, semantics, and spatial hierarchy within 3DSGs for static scenes.
Subsequent methods support incremental construction from RGB~\cite{wu2023incremental}, RGB-D~\cite{wu2021scenegraphfusion}, or RGB+LiDAR sequences~\cite{Rosinol21ijrr-Kimera}, enabling online construction and dynamic updates.
Leveraging foundation models, ConceptGraphs~\cite{gu2024conceptgraphs} and HOV-SG~\cite{Werby-RSS-24} integrate open-vocabulary semantics into scene graphs, demonstrating applicability to language-prompted navigation.
Owing to their structural compactness and rich semantics, 3DSGs increasingly serve as a bridge between scene understanding and high-level task planning~\cite{pmlr-v164-agia22a, booker2024embodiedragdynamic3dscene, ray2024taskmotionplanninghierarchical,  Yan2025Ral3dsgManipulation, Maggio2024Clio}, with broad adoption in embodied intelligence~\cite{Honerkamp2024RAL, hou2025hidynagraphhierarchicaldynamic, yinneurips2024_SG_Nav, Greve2024ICRA_CollaborativeDriving, wang2025curiousbotinteractivemobileexploration}.
However, these methods address semantic and spatial reasoning exclusively, leaving functional interactions unmodeled.

\paragraph{Functional Scene Understanding.}
This area moves beyond “what” and “where” to capture “how” objects interrelate.
IFR-Explore~\cite{li2022ifrexplore} learns inter-object functional relations in synthetic environments but only predicts whether a relation exists, without specifying its type. As it relies on ground-truth 3D data for inference, its real-world applicability is limited. To support functional understanding in real scenes, several benchmarks and methods have emerged.
SceneFun3D~\cite{delitzas2024scenefun3d} provides fine-grained annotations for functionality and affordances of interactive parts.
OpenFunGraph~\cite{zhang2025open} and FunGraph~\cite{rotondi2025fungraph} propose open-vocabulary functional 3D scene graphs that model object-part and object-object interactions using LLMs and 2D vision-language models.
Other works~\cite{engelbracht2025spotlight} explore interaction-driven scene graphs and affordance detection for a distinct set of objects (``lamp", ``switch", ``handle").

Our work builds on this direction but introduces a key advancement: a factor graph framework that jointly reasons over all functional edges by integrating LLM-derived priors and geometric cues. This enables holistic inference with better-calibrated edge confidence predictions, which is critical for informed planning and decision-making in real-world deployments.

\begin{figure*}[t]
    \centering
    \vspace{-4mm}
    \includegraphics[width=0.95\linewidth]{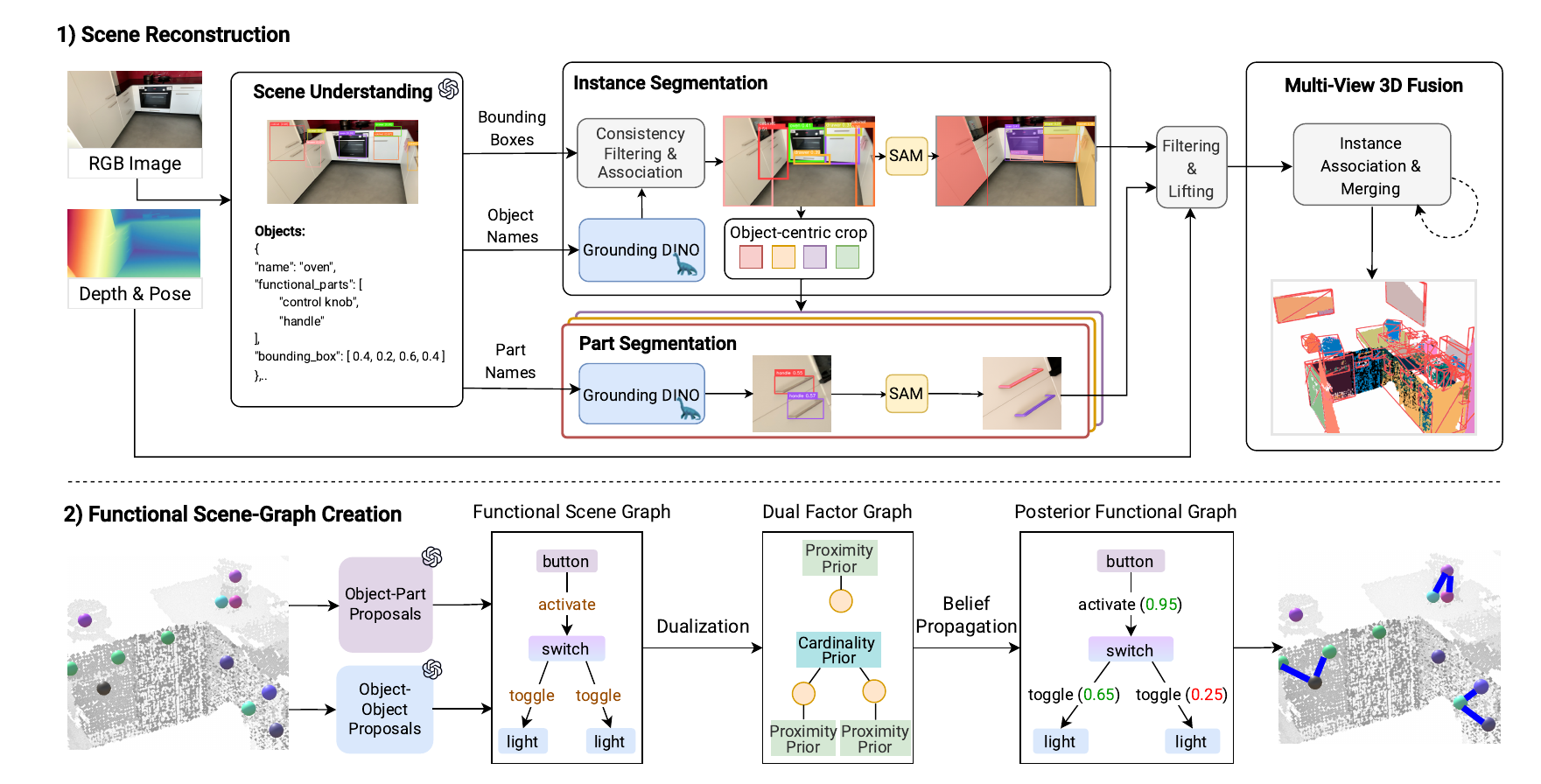}
    \caption{\textbf{Overview of \funfact:} Given Posed RGB-D images \funfact generates scene reconstructions and functional 3D scene graphs in two stages: i) \emph{Scene Reconstruction.} Given a set of RGB-D images and respective poses, we extract bounding boxes, scene description, object list, and candidate part names using \texttt{GPT-4.1}. These textual cues are used to obtain open-vocabulary object detections with GroundingDINO, which are filtered for consistency and turned into region proposals and SAM-based instance masks. A second GroundingDINO + SAM branch segments functional parts conditioned on the predicted object and part names. Finally, we lift all object and part instances to 3D and fuse them across views yielding a coherent, part-aware 3D reconstruction that forms the basis for the subsequent functional scene graph inference.
    ii) \emph{Functional Scene-Graph Creation.} Given the semantic 3D reconstruction and the part / object nodes, \texttt{GPT-4.1} proposes object–object and object–part relations to form an initial functional scene graph. We convert this graph into a dual factor graph with different priors and perform belief propagation to obtain calibrated edge probabilities. This yields the posterior functional scene graph grounded in the reconstructed scene. \vspace{-5mm}}
    \label{fig:overview}
\end{figure*}

\section{Method}

To infer geometric and functional structure from potentially ambiguous and incomplete visual evidence, we design a unified pipeline that combines object- and part-centric scene reconstruction with a probabilistic factor-graph formulation for functional reasoning. Our method grounds open-vocabulary object and part proposals from foundation models into a consistent 3D representation, forming the structural backbone for downstream inference. We then translate semantically plausible functional hypotheses into candidate relations and jointly reason over them through a dual factor graph.
An overview is shown in \cref{fig:overview}.

\subsection{Scene Reconstruction}
Given posed RGB-D observations, \funfact first builds an explicit 3D object- and part-centric representation, which then serves as the backbone for our functional scene graph.

\paragraph{VLM-based object and part proposals.}
We begin by querying a large vision-language model (VLM) with the current RGB image\footnote{We employ \texttt{gpt-4.1-2025-04-14}.}. The VLM is prompted to detect functional objects (e.g., \emph{coffee machine, cabinet, trash can}) and, for each object, output a hierarchical semantic label covering both the object and its functional parts (e.g., \emph{button, handle, pedal}), a concise object-level description, and a coarse 2D bounding box in normalized \texttt{xyxy} format (with coordinates scaled to [0, 1] relative to image dimensions). This step identifies salient functional components and provides semantic and spatial cues for detection.

\paragraph{Object detection and hallucination filtering.}
VLMs can hallucinate non-existent or implausible parts. Furthermore, their predicted bounding boxes tend to be spatially noisy.
To robustly ground the proposed objects, we run GroundingDINO~\cite{liu2024grounding} on the original image using the VLM-proposed object labels as queries.
We then cross-check the detections with the coarse VLM bounding boxes, forming an ensemble of models: proposals that are not detected by GroundingDINO or that strongly disagree with the VLM's coarse bounding boxes are discarded.
This filtering step suppresses hallucinated objects and associates VLM proposals with their grounded detections.

\paragraph{Part-centric detection within object crops.}
For each validated object, we expand its fine bounding box predicted by GroundingDINO, crop the image around the box, and run GroundingDINO again using the VLM-proposed part names of this object as textual queries (e.g., \emph{handle}, \emph{button}).
Cropping focuses the detector on a smaller region, effectively increasing the relative resolution of the object and its parts, which improves the detector's ability to localize small, fine-grained functional components.

\paragraph{Part filtering and consistency checks.}
The raw part detections are further refined by geometric and structural filtering.
We discard parts whose bounding boxes are too large or too small relative to their parent object, as well as parts that have insufficient overlap with the parent object’s bounding box.
This ensures that retained parts are spatially consistent with their parent objects and removes spurious detections on background or unrelated surfaces. Details for our filtering approach are provided in the Appendix (\cref{app:filtering}).
\newpage

\paragraph{Multi-view fusion and object–parts graph construction.}
Following BBQ~\cite{linok2024barequeriesopenvocabularyobject}, we back-project all validated object and part detections into 3D and fuse them across views to obtain consolidated object- and part-level point clouds.
Objects that are consistently observed to overlap across multiple views are merged into a single 3D instance; in such cases, the parts associated with the individual objects are inherited by the merged object.
The resulting representation is an object- and part-centric 3D map together with an explicit graph that encodes object–parts relations.

\subsection{Functional Scene-Graph Creation}

\begin{figure}
    \centering
    \includegraphics[width=0.9\linewidth]{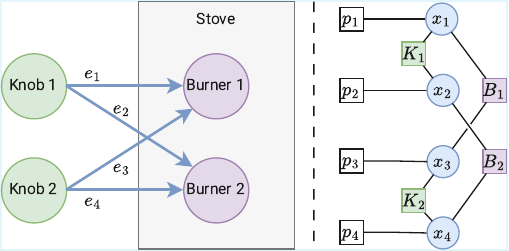}
    \caption{\textbf{Functional Scene Graph and its Dual Factor Graph.}
    \textit{Left:} Candidate functional scene graph with edges $e_1,\dots,e_4$ representing knob–burner relations.
    \textit{Right:} The dual factor graph, where each binary variable $x_i$ \legendcircle{ReneBlue} is the dual of edge $e_i$, encoding whether that relation is present.
    \legendbox{white} $p_i$: unary proximity prior on $x_i$;
    \legendbox{objectgreen} $K_i$: cardinality factor enforcing one-to-one association per knob;
    \legendbox{objectpurple} $B_i$: cardinality factor enforcing one-to-one association per burner.
    }
    \label{fig:factorgraph}
\end{figure}

Given the object- and part-centric 3D map, \funfact constructs candidate functional relations between nodes (e.g., \emph{knob}--\emph{burner}, \emph{handle}--\emph{door}) and encodes them as binary variables constrained by prior factors in a dual functional factor graph, as illustrated in \cref{fig:factorgraph}. The ``dual'' nature of this graph stems from an inversion of the original scene graph structure: nodes from the scene graph become edge constraints (factors) in the factor graph, while scene graph edges (the candidate relations) become the variables.

\paragraph{LLM-based functional relation priors.}
For every object with functional parts, we query an LLM with the object's label, description, and the labels of all its parts (e.g., \textit{knob}, \textit{burner}, \textit{door handle}), asking it to propose plausible semantic \emph{functional relations} among them (e.g., \textit{knob \underline{controls} burner}, \textit{handle \underline{opens} door}).
For each proposed relation type, the LLM also predicts whether it is typically \emph{one-to-one} (e.g., each burner has a dedicated knob) or follows a more flexible cardinality pattern (e.g., a control panel with many buttons for one device). For a scene with $N$ objects, we represent the full set of proposed relation templates as $\mathcal{T} = \{\mathcal{T}_k\}_{k=1}^N$, where $\mathcal{T}_k = \{r_{k,j}\}_{j=1}^{M_k}$ is the set of $M_k$ relation templates proposed for the $k$-th object.

\paragraph{Dual factor graph construction.}
For each relation template $r_{k,j}$, we enumerate all part--object and part--part combinations that match the semantic types of the template (e.g., all pairs of \textit{stove knob} and \textit{stove burner} on a stove) and connect them with candidate functional edges. Let $\mathcal{E}_{k,j} = \{e_i^{k,j}\}_{i=1}^{E_{k,j}}$ denote the set of candidate edges instantiated for $r_{k,j}$, where $E_{k,j}$ is the total number of edges resulting from this exhaustive match. From these edges, we construct a local factor graph with variables $\mathcal{X}_{k,j}=\{x_i^{k,j}\}_{i=1}^{E_{k,j}}$. Each binary variable $x_i^{k,j} \in \{0,1\}$ is the dual of edge $e_i^{k,j}$, indicating whether that functional edge is present.
This construction yields a densely connected local functional group (e.g., a complete bipartite graph between all knobs and burners on a stove), which we subsequently disambiguate through probabilistic inference. \\
\underline{Cardinality-based constraint factors.}
To encode structural priors such as one-to-one or one-to-many mappings, we introduce \emph{cardinality factors} $\phi_{\text{card}}(\cdot)$ over variables within each local group. For one-to-one relations, these factors penalize configurations where a single part connects to multiple counterparts (e.g., one knob controlling several burners, or one burner controlled by several knobs) or where a part is not connected to any counterpart. Concretely, for a part node $n$ (e.g., a specific knob) involved in one-to-one relations, let $\mathcal{X}_n$ denote the variables whose dual edges are incident to $n$, and let $d_n = \sum_{x \in \mathcal{X}_n} x$ be the number of active connections. We define the cardinality factor as:
$$
\phi_{\text{card}}(\mathcal{X}_n) =
\begin{cases}
b^{d_n - 1} & \text{if } d_n \geq 1, \\
b^2 & \text{if } d_n = 0,
\end{cases}
$$
where $b \in (0,1)$ is a hyperparameter controlling the strength of the penalty. Intuitively, this factor imposes soft constraints on the number of active edges incident to the node, making configurations with too many connections (or zero connections) less likely under the model, thereby favoring structurally plausible assignments.\\
\underline{Proximity-based prior factors.}
Finally, for each variable $x_i^{k,j}$, we assign a prior belief based on the length of its dual edge $e_i^{k,j}$ (i.e., the Euclidean distance between the nodes it connects). We encode these beliefs as unary proximity factors:
\begin{equation}
    \phi_{\text{prox}}(x_i^{k,j})= e^{-\frac{d(e_i^{k,j})}{\lambda_{k,j}}},
\end{equation}
where $d(e_i^{k,j})$ is the length of the edge $e_i^{k,j}$ and $\lambda_{k,j}$ is a scaling parameter defined as the median length of all edges in the local candidate edge set $\mathcal{E}_{k,j}$.
This formulation biases the graph toward selecting the closer connection while still allowing the factor graph to correct mistakes using the cardinality constraints.
The result is a set of local functional factor graphs where each candidate relation is represented as a binary variable, regularized by both proximity priors and structured cardinality constraints, ready for holistic inference in the global \funfact model. \newpage

\paragraph{Object--Object functional proposal.}
Analogous to the part--object and part--part proposals in the previous section, we use the LLM to suggest plausible inter-object functional relations (e.g., \emph{sponge cleans countertop}, \emph{knife can cut apple}), along with their typical cardinality patterns. But for object-object relations, we do not enforce the proximity prior by default, and instead instruct the LLM to suggest which relations require proximity (e.g., curtains cover windows). We then instantiate local factor graphs over all edges that are either ``one-to-one'' or require proximity, or both, equip them with the same cardinality-based constraint factors and proximity prior factors, and jointly optimize these object–object edges together with the part-centric relations in the global \funfact model.

\paragraph{Probabilistic inference via belief propagation.}
We implement the dual functional factor graph using pgmpy \cite{Ankan2024} and perform belief propagation to infer the joint distribution over all candidate functional edges. To accelerate inference, we identify disjoint connected components, denoted as $\mathcal{C}_m (m=1,2,..., M)$, which are isolated subgraphs that do not share prior or constraint factors (e.g., knowing a knob controls a burner does not help disambiguate connections between remote controls and TVs), and run inference on each component separately. For a given component $\mathcal{C}_m$ with variable set $\mathcal{X}_m$, the joint distribution is $P(\mathcal{X}_m) = \frac{1}{Z_m} \prod_{x \in \mathcal{X}_m} \phi_{\text{prox}}(x) \prod_{f \in \mathcal{F}_m} \phi_{\text{card}}(\partial f)$, where $\mathcal{F}_m$ denotes the cardinality factors in $\mathcal{C}_m$, $\partial f \subseteq \mathcal{X}_m$ the variables connected to factor $f$, and $Z_m$ a normalization constant. After convergence, we marginalize this distribution to obtain per-edge confidence scores, which are thresholded to produce the final functional scene graph.
\section{Results}
\begin{figure}
    \centering
    \includegraphics[width=1.0\linewidth]{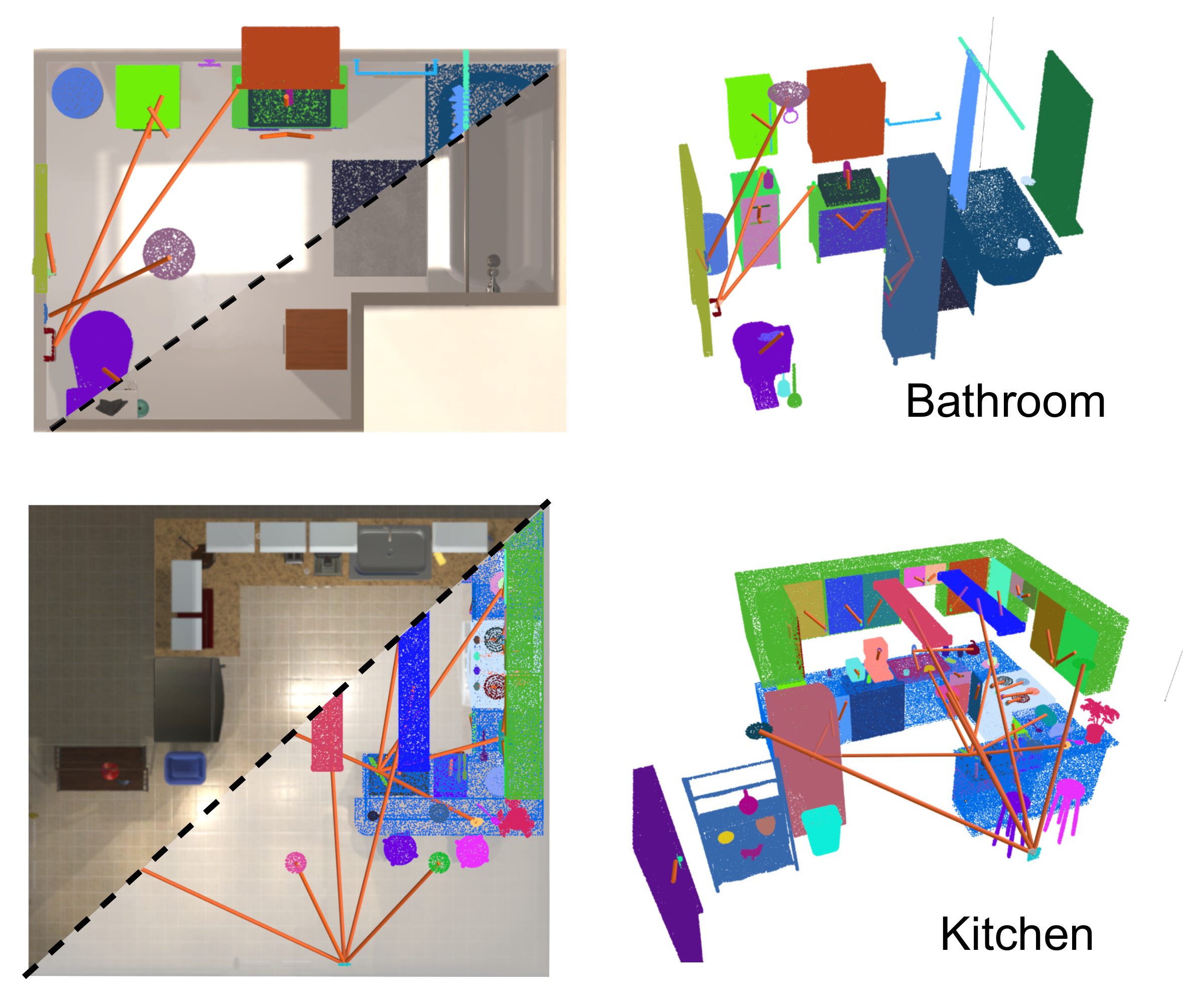}
\caption{\textbf{Examples of two of our newly annotated AI2-THOR environments.} For two scenes (bathroom, top; kitchen, bottom) we show a top-down view with mapped functional objects (left) and the corresponding instance-, part-segmentation and functional edges (right), illustrating our functional annotations. \vspace{-5mm}}
\label{fig:ai2thor_envs}
\end{figure}

\begin{figure}
    \centering
    \includegraphics[width=1.0\linewidth]{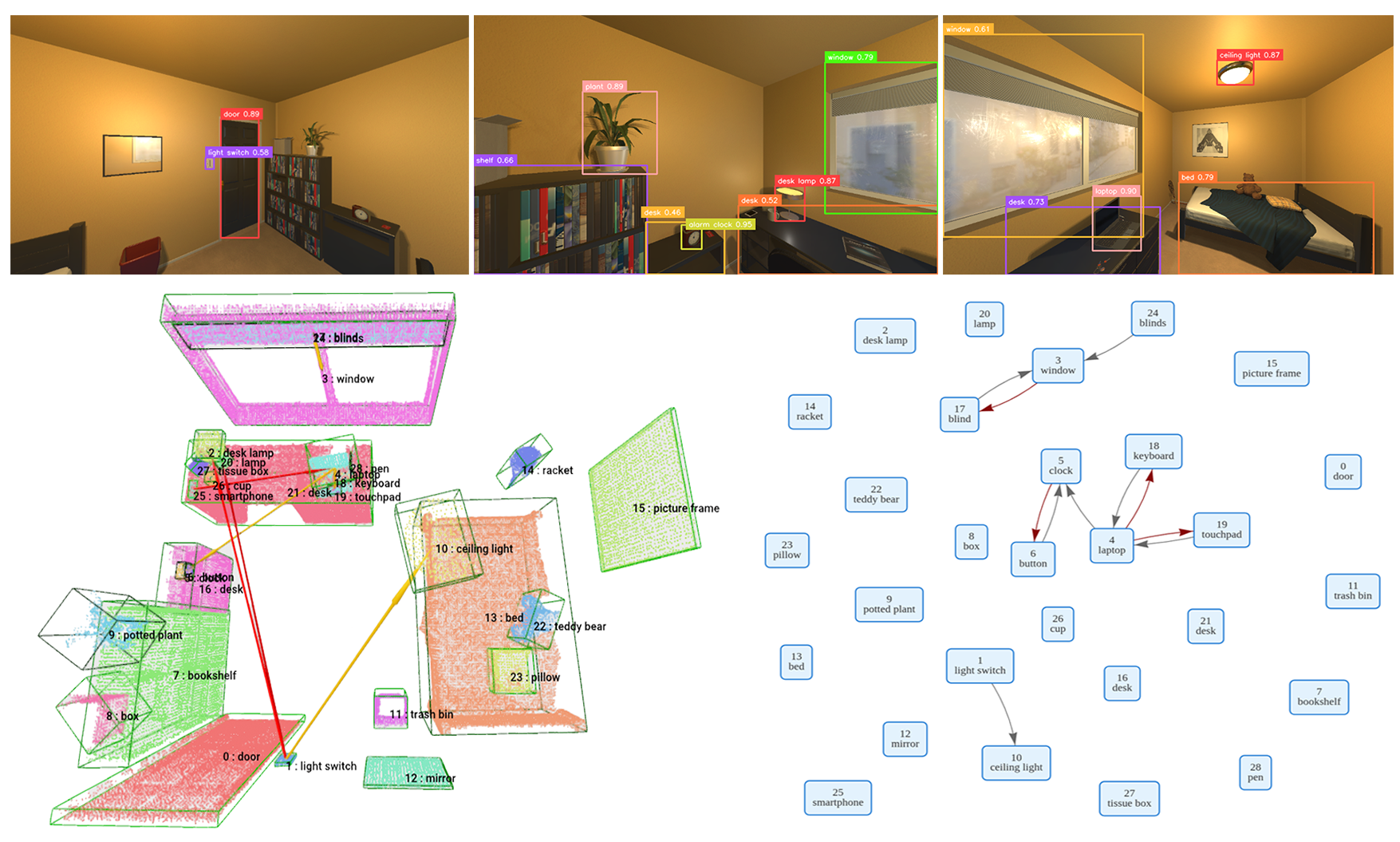}
\caption{\textbf{Qualitative Results.} \textit{Top:} Input images with detected functional objects. \textit{Bottom-left:} Reconstructed object and part point clouds with predicted functional relations (red: confidence $<$ 0.5; yellow: confidence $\geq$ 0.5). \textit{Bottom-right:} Final functional 3D scene graph after confidence thresholding; red edges indicate object-part hierarchy and gray edges indicate functional relations.
\vspace{-7mm}}
\label{fig:qualitative_main}
\end{figure}

We evaluate our \funfact across both real-world datasets and our newly introduced \funthor benchmark to assess its effectiveness in reconstructing functional scene graphs and predicting functional relations with better-calibrated confidence estimates.
The experiments evaluate mapping performance, functional edge quality, and confidence calibration, with comparisons to state-of-the-art baselines.
Fig.~\ref{fig:qualitative_main} shows qualitative results of \funfact. Our pipeline detects fine-grained functional parts (e.g., the keyboard and touchpad of a laptop) and predicts both intra-object relations (e.g., keyboard inputs commands to laptop) and inter-object relations (e.g., light switch turns on/off ceiling light).

\subsection{Functional AI2-THOR (\funthor)}
Existing datasets for functional scene understanding, such as SceneFun3D~\cite{delitzas2024scenefun3d} and FunGraph3D~\cite{zhang2025open}, have been instrumental in driving progress, but remain limited in scope. SceneFun3D primarily focuses on part-level affordances (e.g., knobs,  handles), with limited coverage of inter-object relations, while FunGraph3D extends to inter-object functionality but suffers from sparse and partially heuristic annotations. As a result, these datasets provide incomplete coverage of functionally important entities and relations, limiting fair evaluation of precision and calibration.

To address these limitations, we introduce \emph{Functional AI2-THOR (\funthor)}, a new synthetic benchmark built on top of the AI2-THOR~\cite{kolve2017ai2} scenes. \funthor contains 12 scenes
spanning 4 environment types (i.e., kitchen, living room, bedroom, and bathroom; three each),  with part-level geometry and 26 types of functional relations inherently supported by the simulator.
Each scene includes 60 uniformly sampled, posed RGB-D frames and complete ground-truth annotations for nodes and functional relations, spanning a total of 720 images.
All functional annotations are automatically generated based on object properties and affordances natively supported in AI2-THOR, ensuring comprehensive coverage of all functionally relevant entities and relations in each scene.
\cref{fig:ai2thor_envs} shows example scenes and their annotations from \funthor. Further details are provided in the Supplementary material.

\subsection{Mapping Performance}
\label{subsec:mapping}

\begin{table*}[!t]
\centering
\caption{\textbf{Scene Reconstruction Comparison.}  We report Recall@K (\textbf{R@3}/\textbf{R@10}; higher is better) for three node categories—\emph{Objects}, \emph{Interactive Elements}, and \emph{Overall Nodes}—on \textbf{SceneFun3D} and \textbf{FunGraph3D}.
Bold numbers mark the best result per column.
Compared to Open3DSG~\cite{koch-2024-open3dsg}, ConceptGraph~\cite{gu2024conceptgraphs}, ConceptGraph variant, and OpenFunGraph~\cite{zhang2025open}, \textbf{FunFact} attains the highest recall across nearly all categories and both datasets, indicating more reliable recovery of functional entities and relations.}
\renewcommand{\arraystretch}{1.1}
\resizebox{1.0\textwidth}{!}{
\begin{tabular}{lcccccc|cccccc}
\toprule
\textbf{Methods} &
\multicolumn{6}{c|}{\textbf{SceneFun3D}} &
\multicolumn{6}{c}{\textbf{FunGraph3D }} \\
\cmidrule(lr){2-7}\cmidrule(lr){8-13}
& \multicolumn{2}{c}{Objects $(\uparrow)$ } & \multicolumn{2}{c}{Inter.\ Elements $(\uparrow)$ } & \multicolumn{2}{c|}{Overall Nodes $(\uparrow)$ }
& \multicolumn{2}{c}{Objects $(\uparrow)$ } & \multicolumn{2}{c}{Inter.\ Elements $(\uparrow)$ } & \multicolumn{2}{c}{Overall Nodes $(\uparrow)$ }
 \\
& R@3 & R@10 & R@3 & R@10 & R@3 & R@10
& R@3 & R@10 & R@3 & R@10 & R@3 & R@10
 \\
\midrule
Open3DSG           & 61.2 & 70.7 & 54.4 & 61.8 & 56.7 & 64.7 & 50.9 & 58.1 & 21.8 & 33.9 & 33.4 & 43.6  \\
ConceptGraph     & 71.3 & 77.1 &  6.6 &  8.6 & 28.3 & 31.4 & 58.0 & 66.3 &  2.5 &  4.1 & 20.1 & 25.2 \\
ConceptGraph + IED & 71.3 & 77.1 & 53.1 & 59.5 & 60.1 & 66.0 & 58.0 & 66.3 & 20.5 & 33.4 & 38.9 & 45.0 \\
OpenFunGraph       & {81.8} & {87.8} & \textbf{71.0} & \textbf{79.5} & {73.0} & {82.8}
                         & {70.7} & {79.1} & {44.4} & {57.6} & {55.5} & {65.8}
            \\
\midrule

 FunFact (Ours)      & \textbf{90.5} & \textbf{93.3} &
{64.6} & {78.8}
& \textbf{73.2} & \textbf{83.6}&

\textbf{91.1} & \textbf{96.6} &
\textbf{68.3} & \textbf{78.7} &
\textbf{77.9} & \textbf{86.2} \\

\bottomrule
\end{tabular}
}
\label{tab:reconstruction}
\vspace{-6pt}
\end{table*}

We evaluate mapping quality on SceneFun3D and FunGraph3D following the Recall@K protocol introduced in OpenFunGraph~\cite{zhang2025open}.
For each ground-truth node, we loop over all predicted nodes and find the first one with a non-zero 3D bounding box IoU. We then compute the CLIP-based \cite{radford2021learning} cosine similarities between the predicted textual label and all labels in the dataset, and treat the prediction as a match if the ground-truth label ranks among the top-K most similar labels (K=3, 10).

As shown in Tab.~\ref{tab:reconstruction}, \funfact achieves consistently higher recall across object, and overall node categories compared to all baselines. On SceneFun3D, our model maintains competitive recall despite annotation biases that label visually distinct interactive elements (e.g., ``pedal'' or ``drain'') under generic classes such as ``handle'' or ``knob'', as shown in the Appendix. This coarse labeling lowers open-vocabulary alignment scores, and the larger performance gap between K=3 and K=10 further highlights this issue, as our fine-grained predictions are incorrectly penalized when matched against these generic annotations.

On FunGraph3D, where annotations are more specific, \funfact outperforms OpenFunGraph by significant margins in both Recall@3 and Recall@10, particularly for small interactive elements. This improvement stems from our hierarchical object-part mapping pipeline, which enables robust detection and fusion of fine-grained components often missed by flat object-centric baselines.

\subsection{Functional Relationships}

\begin{table*}[ht!]
\centering
\caption{\textbf{Triplet Evaluation.} We report node association, edge prediction and overall triplet recall as  Recall@K (R@5/R@10) on SceneFun3D and FunGraph3D and (R@3/R@5) on \funthor.}
\renewcommand{\arraystretch}{1.1}
\resizebox{\textwidth}{!}{
\begin{tabular}{lcccccc|cccccc|cccccc}
\toprule
\textbf{Methods} &
\multicolumn{6}{c|}{\textbf{SceneFun3D}} &
\multicolumn{6}{c|}{\textbf{FunGraph3D }} &
\multicolumn{6}{c}{\textbf{FunThor}} \\
\cmidrule(lr){2-7}\cmidrule(lr){8-13}\cmidrule(l){14-19}
& \multicolumn{2}{c}{Node Assoc. ($\uparrow$)} & \multicolumn{2}{c}{Edge Pred. ($\uparrow$)} & \multicolumn{2}{c|}{Overall Triplets ($\uparrow$)}
& \multicolumn{2}{c}{Node Assoc. ($\uparrow$)} & \multicolumn{2}{c}{Edge Pred. ($\uparrow$)} & \multicolumn{2}{c|}{Overall Triplets ($\uparrow$)}
& \multicolumn{2}{c}{Node Assoc. ($\uparrow$)} & \multicolumn{2}{c}{Edge Pred. ($\uparrow$)} & \multicolumn{2}{c}{Overall Triplets ($\uparrow$)} \\
& R@5 & R@10 & R@5 & R@10 & R@5 & R@10
& R@5 & R@10 & R@5 & R@10 & R@5 & R@10
& R@3 & R@5 & R@3 & R@5 & R@3 & R@5 \\
\midrule

OpenFunGraph       & \textbf{68.3} & \textbf{73.0} & \textbf{88.1} & \textbf{96.2} & \textbf{60.4}
                         & \textbf{70.3} & {45.8} & {49.3} & {65.1} & \textbf{91.4} & {29.8}
                         & {45.0}   & {26.4}   & {26.4}   & {59.0}   & {66.7}   & {15.1} & {17.6} \\
\midrule

 FunFact (Ours)      & {62.1} & {71.8} &
{66.1} & {80.7}
& {41.0} & {57.9}&

\textbf{71.1} & \textbf{80.0} &
\textbf{67.9} & {79.9} &
\textbf{48.7} & \textbf{63.9} &

 \textbf{60.8}   & \textbf{62.8}   &
 \textbf{88.9}   & \textbf{87.1}   &
 \textbf{54.1}   & \textbf{54.7} \\

\bottomrule
\end{tabular}
}
\label{tab:fungraph_results}
\vspace{-1.2em}
\end{table*}
Following the evaluation protocol of OpenFunGraph, we assess functional relation prediction using the Recall@K metric on SceneFun3D, FunGraph3D, and our newly proposed \funthor dataset. A functional relation is represented as a triplet (subject, interaction, object), and a predicted triplet is considered correct only if both its nodes and interaction label match the ground truth. For node-level matching, the predicted subject and object must each match the ground-truth entities spatially and semantically, as described in Sec.~\ref{subsec:mapping}. The interaction label is evaluated only when both node matches are confirmed: we compute BERT~\cite{devlin2019bert} similarities between the predicted relation label and all ground-truth relation labels of the dataset, and the prediction is correct if the ground-truth label ranks within the top K. We report Recall@K for K=5 and 10 on SceneFun3D and FunGraph3D, and K=3 and 5 on \funthor following its denser annotation protocol.

Tab.~\ref{tab:fungraph_results} summarizes our results. We report three metrics: node association recall (i.e., subject-object pairs without considering the interaction label), full triplet recall (i.e., subject, interaction, object), and edge prediction recall (i.e., the rate of matched triplets among all predictions with correct subject-object pairs). On FunGraph3D, \funfact substantially outperforms OpenFunGraph across all metrics except edge prediction R@10. This slight drop is expected: FunFact detects significantly more functional elements, yielding a larger denominator when computing edge prediction recall. Importantly, when considering full triplets (reflecting real functional reasoning rather than isolated edges), \funfact achieves notably higher recall, demonstrating the value of jointly modeling relation structure instead of predicting edges independently.

On \funthor, which provides exhaustive annotations across both object-object and part-object relations, \funfact again achieves large improvements over OpenFunGraph. The gains of {39.0pp}\,/\,{37.1pp} (R@3/5)  in overall triplet recall show the benefits of hierarchical relation proposal and our factor-graph inference, which resolves visually ambiguous cases that pairwise baselines misclassify.

Performance on SceneFun3D follows a different trend. Here, OpenFunGraph attains higher recall with respect to SceneFun3D's highly generic labels such as ``handle'', ``knob'', or ``button''. These broad categories lead to systematic mismatches with \funfact's open-vocabulary predictions (e.g., ``television stand'' vs ``cabinet''), despite the predictions being semantically correct. We identified several such cases manually. In contrast, \funthor, via its rule-based fine-grained annotation, exhibits far fewer label-mismatch artifacts. This leads to a cleaner picture of functional prediction accuracy and highlights a limitation of CLIP/BERT-based matching protocols when interacting with open-vocabulary outputs.

Overall, the triplet results across all three datasets demonstrate that \funfact's holistic factor-graph formulation and hierarchical relation proposal meaningfully improve functional scene understanding, particularly in settings with dense annotations and diverse relation types.

\subsection{Detailed Confidence Evaluation}
\label{subs:aithor}

In prior works, functional relation prediction performance has only been evaluated using Recall@K metrics due to the sparsity of the manually annotated edges.
Thanks to the dense functional relation annotations in \funthor, we are able to comprehensively evaluate not only recall but also precision and confidence calibration of predicted functional relations. In particular, we report the expected calibration error (ECE)~\cite{guo2017calibration} for edge confidence. This measures the error with respect to a perfect confidence where, e.g., out of all 60\% confident edges, 60\% are correct. We report the ECE over all edges, as well as specifically for challenging ambiguous cases of light switches and stove knobs.

We threshold predicted functional relation probabilities at 0.5 for precision and recall, and use all predictions to compute ECE. Since OpenFunGraph reports confidence only for edges involving ``outlet,'' ``switch,'' and ``remote,'' we assign confidence 1.0 to all other predicted relations. Other baselines do not report confidence scores and are excluded from this evaluation.

For each metric, we report top-3 retrieval for object matching and relation matching. We calculate the overall triplet precision as $P_{tr}={n_{ma}}/{n_{de}}$ where $n_{ma}$ is the number of correctly predicted functional relation triplets and $n_{de}$ is the total number of predicted triplets. The overall triplet recall is calculated as $R_{tr}={n_{ma}}/{n_{gt}}$ where $n_{gt}$ is the total number of ground truth triplets in the scene. ECE is computed following the standard definition~\cite{guo2017calibration}:
\begin{equation}
    ECE = \sum_{m=1}^{M} \frac{|B_m|}{n} \left| \text{acc}(B_m) - \text{conf}(B_m) \right|,
\end{equation}
where $n$ is the total number of predictions, $M$ is the number of bins, $B_m$ is the set of samples whose predicted confidences fall into bin $m$, $\text{acc}(B_m)$ and $\text{conf}(B_m)$ are the accuracy and average confidence of bin $m$ respectively. We use 4 bins ($M = 4$) for this evaluation.

According to the results in \cref{tab:fungraph_results_sm}, our method outperforms OpenFunGraph across all metrics, demonstrating that our method not only predicts more accurate functional relations but also provides better calibrated confidence scores.
Notably, our method achieves a substantial 8.5\% absolute improvement in precision compared to OpenFunGraph, highlighting its effectiveness in resolving visual ambiguities through scene-wide context.

\begin{table*}[t!]
\centering
\caption{\textsc{Ablation study and calibration scores on the \funthor Dataset.}
We report mapping quality (Recall@3), functional edge scores (Precision, Recall, F1; higher is better), and calibration metrics (ECE, ECE-ambiguous; lower is better).}
\renewcommand{\arraystretch}{1.1}
\resizebox{0.95\textwidth}{!}{
\begin{tabular}{lccc|ccc|cc}
\toprule
\multirow{3}{*}{\textbf{Methods}} &  \multicolumn{3}{c}{\textbf{Mapping}}  & \multicolumn{5}{c}{\textbf{Functional Graph}}  \\
\cmidrule(lr){2-4} \cmidrule(lr){5-8} \cmidrule(lr){8-9}
& \multicolumn{3}{c|}{Recall @ 3 $(\uparrow)$} & \multicolumn{1}{c}{Prec. [\%]} & \multicolumn{1}{c}{Recall [\%]} & \multicolumn{1}{c|}{F1 [\%]} & \multicolumn{2}{c}{ECE} \\
& \multicolumn{1}{c}{Objects} & \multicolumn{1}{c}{Inter.\ Elem.} & \multicolumn{1}{c|}{Overall Nodes} & $(\uparrow)$ & $(\uparrow)$ & $(\uparrow)$  & \ \  \ \ \ \  All $(\downarrow)$ \ \  \ \ \ \  & Ambiguous $(\downarrow)$ \\
\midrule
OpenFunGraph     & 54.6 & 41.1 & 51.2 & 23.4 & 12.2 & 16.0 & 0.43 & 0.51 \\
\midrule
FunFact (Ours)                                    & 68.2 & \textbf{69.5} & \textbf{68.5} & \textbf{31.9} & 49.3 & \textbf{38.7} & \textbf{0.36} & \textbf{0.07} \\
\quad w/o FactorGraph                             & 68.2 & \textbf{69.5} & \textbf{68.5} & 21.9 & \textbf{53.4} & 31.1 & 0.70 & 0.45 \\
\quad w/o Hierarchical Object and Part proposal    & \textbf{68.9} & 41.8 & 62.1 & 21.6 & 18.2 & 19.8 &  \textbf{0.36} & 0.14 \\
\bottomrule
\end{tabular}
}
\label{tab:fungraph_results_sm}
\vspace{-5mm}
\end{table*}

\subsection{Ablation Studies}
To evaluate the effectiveness of different components in our method, i.e., hierarchical object and part mapping and factor graph reasoning, we conduct ablation studies on \funthor. Specifically, we investigate the impact of removing (i) the factor graph reasoning module and (ii) the hierarchical representation of objects and parts during mapping and merging. We follow the same evaluation protocol as in Sec.~\ref{subs:aithor}, using top-3 retrieval for object and relation matching, and filtering out all edges with a confidence score below 0.5.

As shown in Tab.~\ref{tab:fungraph_results_sm}, both components contribute significantly to the overall functional prediction performance. The factor graph reasoning module infers confidence scores for visually ambiguous relations by leveraging scene-wide context.
While this suppresses some uncertain predictions and slightly reduces recall, it substantially improves precision by eliminating low-confidence edges.
The net effect is a marked improvement in F1 score, demonstrating that the module effectively resolves relational ambiguities by reasoning over the full scene.

The hierarchical object-part representation during mapping is equally important. We ablate this component by treating all detected parts as individual objects during mapping and merging. While object mapping recall remains similar to the full model, since the object-type pipeline is unchanged, interactive element mapping recall drops significantly, causing a substantial decrease in triplet precision, recall, and F1. The flattened pipeline loses the ability to focus on objects with small interactive elements and associate them correctly, which is crucial for accurate functional relation prediction.
We further note that without the Hierarchical Object and Part proposal, the interactive element mapping recall is quite close to that of the OpenFunGraph~\cite{zhang2025open} baseline (41.8\% vs 41.1\%), which also treats all interactive elements as individual objects during mapping. This further demonstrates the importance of hierarchical detection of objects and their parts for effective discovery and association of small interactive elements with their parent objects.

\section{Limitations and Future Work}
\funfact offers a principled method for inferring functional structure through a unified scene- and factor-graph formulation; however, opportunities for improvement remain:

First, we observe that our method can both over- and under-segment objects.
For example, a wall with multiple cabinets may occasionally be fused into one instance. Part segmentation is also often ambiguous: while \emph{FunGraph3D}~\cite{zhang2025open} annotates only a single microwave control panel, our method may instead produce separate parts for each button. This finer granularity is rarely captured in existing datasets, making benchmarking difficult and sensitive to visual and semantic ambiguity.

\begin{figure}[t]
    \centering
    \includegraphics[width=0.95\linewidth]{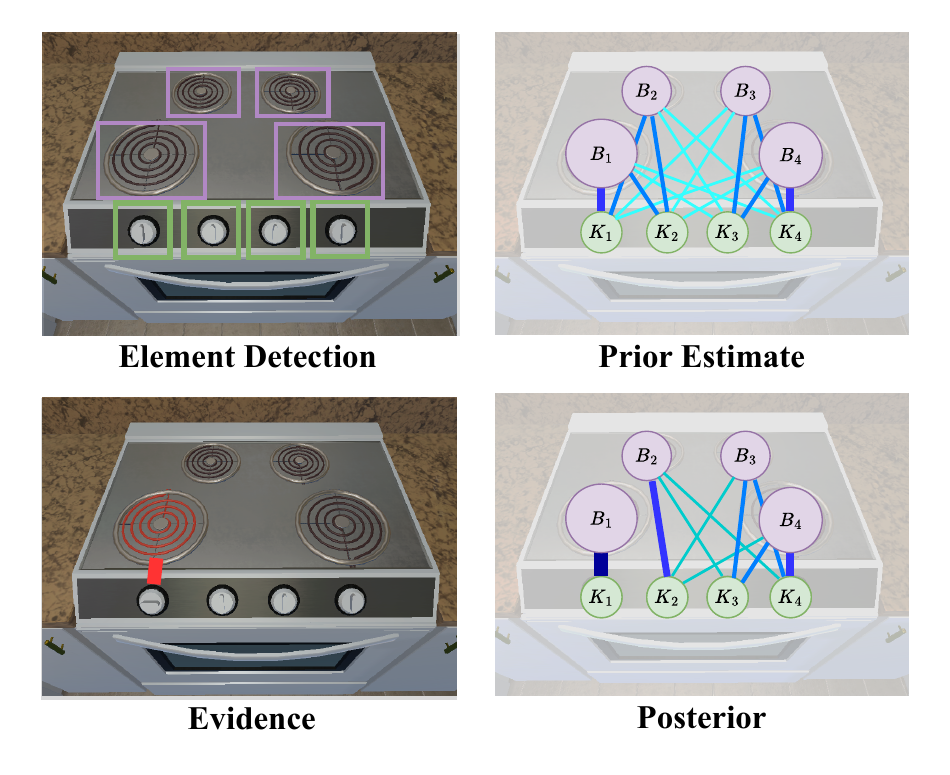}
       \vspace{-5mm}
    \caption{\textbf{Examples of our dual factor graph and interactive verification.} We highlight how our dual graph can be used to calibrate functional predictions over time. Thicker and darker blue edges indicate higher confidence, while thinner and lighter blue edges represent less probable associations. In the beginning, each knob-burner edge receives confidence solely from the proximity and cardinality priors. After introducing evidence (i.e., the left knob controls the left burner plate), our model propagates this information to update the remaining three knob–burner associations.}
    \label{fig:posterior_modeling}
    \vspace{-4mm}
\end{figure}
Second, our pipeline currently relies heavily on LLM-based reasoning, which introduces non-negligible inference latency on the order of several seconds per image.
Although this is still competitive with or faster than some existing SOTA methods, it limits applicability in strict real-time settings and on resource-constrained platforms.
In future work, we aim to bridge the gap to robotics and embodied intelligence more closely. Our formulation naturally incorporates new evidence to update beliefs (as shown in \cref{fig:posterior_modeling}), while better calibration enables more precise identification of uncertain connections. Together, these properties support combining functional scene graph reconstruction with informed robotic verification and real-world exploration.
\section{Conclusion}
We present \funfact, a probabilistic framework for open-vocabulary functional 3D scene understanding. It reconstructs functional objects from posed RGB-D images and proposes semantically plausible relations, which are jointly refined through factor graph inference. By incorporating LLM-derived commonsense priors and geometric cues as factors, \funfact resolves local ambiguities via global scene context, yielding substantially better-calibrated confidence estimates than pairwise reasoning alone.

To enable detailed evaluation of precision and confidence, we introduce \funthor, a synthetic benchmark built on AI2-THOR~\cite{kolve2017ai2} with more systematic and comprehensive functional annotations than existing datasets. Across FunGraph3D~\cite{zhang2025open} and \funthor, \funfact consistently outperforms state-of-the-art baselines, validating the benefits of holistic probabilistic modeling for functional scene understanding.

\section*{Acknowledgements}
This work was partially supported by the ETH AI Center, the Swiss National Science Foundation through the National Centre of Competence in Digital Fabrication (NCCR dfab), and Huawei Tech R\&D (U.K.) through a research funding agreement. Additional support was provided by ETH Foundation Project 2025-FS-352 and SNSF Advanced Grant 216260. The authors also thank Dr.\ Cesar Dario Cadena Lerma for his insightful feedback on the mathematical notation used in this work, and Wanru Zhao for her expert assistance in the preparation of the figures.

{
    \small
    \bibliographystyle{ieeenat_fullname}
    \bibliography{main}
}
\startcontents[supp]
\clearpage
\appendix
\setcounter{page}{1}
\maketitlesupplementary

\begin{center}
\begin{minipage}{0.48\textwidth}
\setlength{\parskip}{0pt}
\makeatletter
\makeatother
{\small
\printcontents[supp]{}{1}{\setcounter{tocdepth}{1}}
}
\end{minipage}
\end{center}
\vspace{5pt}

\section{Part Filtering}
\label{app:filtering}
To filter out spurious part detections on the background or having unreasonably large overlap with the parent objects, we implement a part filter based on the overlap ratio between the part's bounding box and the parent object's bounding box. A part detection is discarded if the intersection between the part's bounding box and the parent object's bounding box occupies:
\begin{enumerate}[label=\arabic*)]
  \item less than 30\% of the part's bounding box (\ie ., background objects are incorrectly detected as functional parts of the parent object), or
  \item more than 70\% of the object's bounding box (\ie, the model misdetects the object itself as one of its functional parts)
\end{enumerate}

\section{\funthor Generation Process}
\label{app:AI2THOR-curation}
To build \funthor, we follow a principled pipeline that produces part-aware geometries and dense functional-relation annotations based on AI2-THOR~\cite{kolve2017ai2} infrastructure. In addition, we generate posed RGB-D images for each scene to support perceptual functional scene understanding.

\begin{figure}[!t]
    \centering
    \begin{subfigure}[b]{0.23\textwidth}
        \centering
        \includegraphics[width=\textwidth]{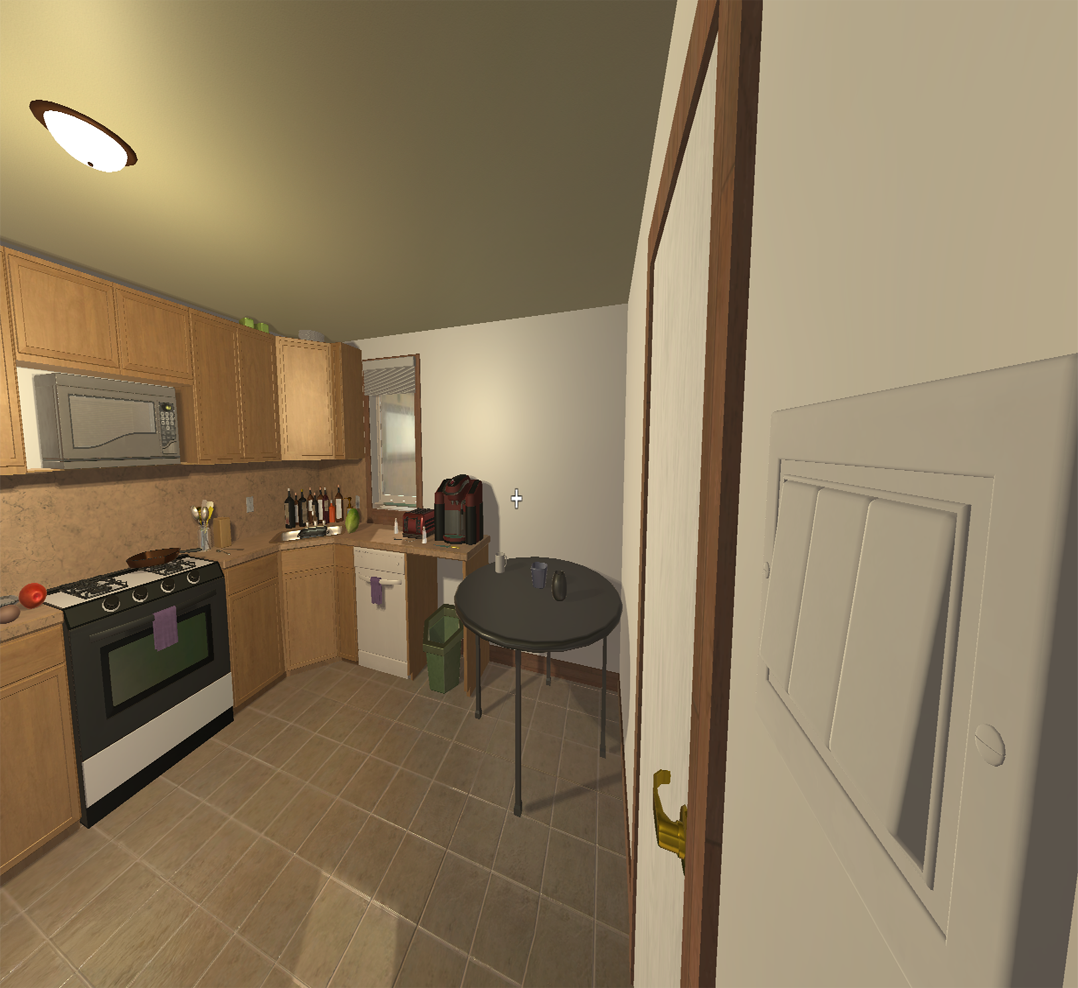}
    \end{subfigure}
    \hfill
    \begin{subfigure}[b]{0.23\textwidth}
        \centering
        \includegraphics[width=\textwidth]{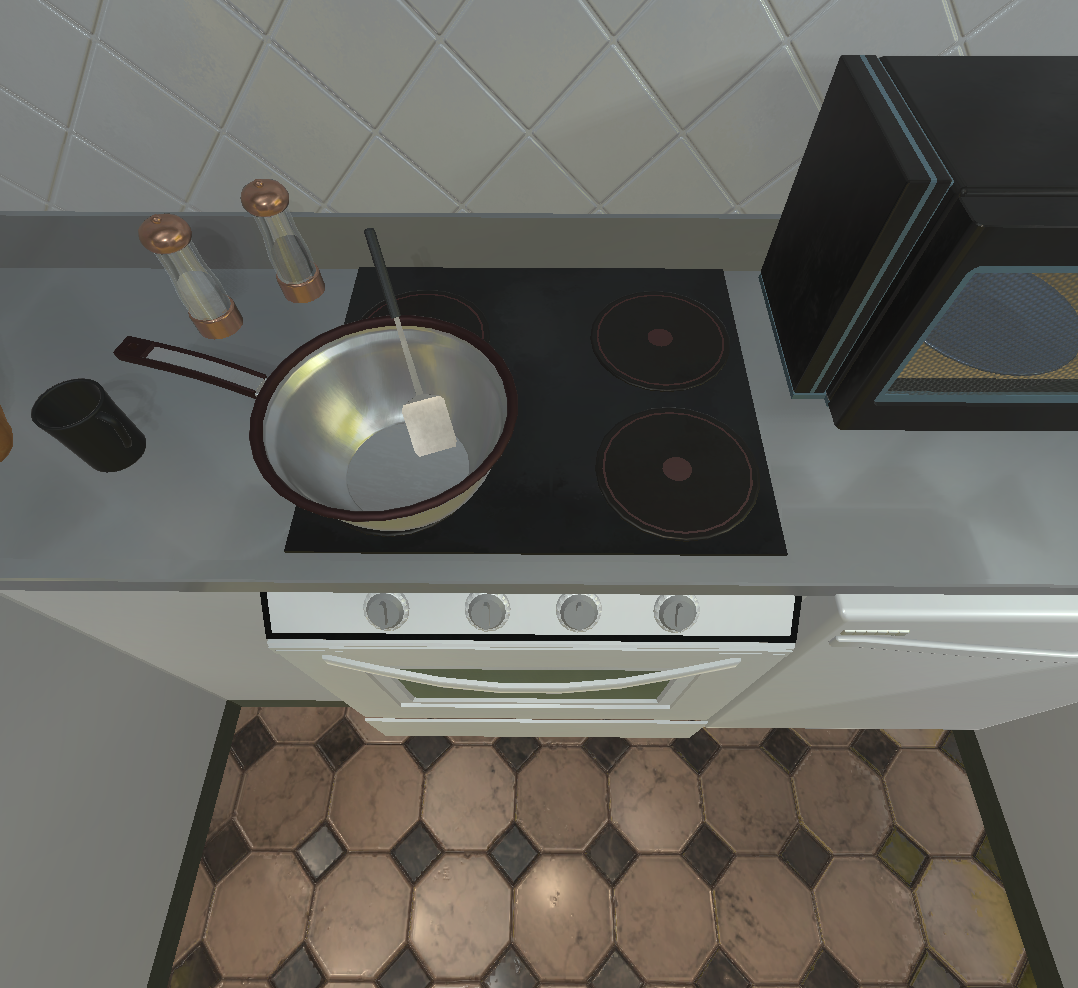}
    \end{subfigure}
    \caption{\textbf{Examples of scene filtering for \funthor.} Scenes are excluded if switch-light associations are ambiguous (left) or if pans are present on stove burners (right). }\label{fig:scene_exclusion}
\end{figure}

\paragraph{Scene selection.}
In all AI2-THOR scenes, a single light switch controls the lighting of the entire environment, regardless of how many light fixtures are present or how many sub-switches the switch includes. To better reflect the real world settings, we exclude scenes where sub-switches cannot be heuristically assigned to controlled lights (\eg, a dual switch controls a single light or a quad switch controls three rows of lights). From the remaining set, we intentionally select scenes with higher visual ambiguity. For example, scenes with a hanging light controlled by a light switch surrounded by multiple floor lamps. For kitchen-type scenes, we additionally filter out those containing a pan on a stove burner to support burner detection. Fig.~\ref{fig:scene_exclusion} illustrates examples of excluded scenes.

\paragraph{Part-level annotation.}
\label{app:part-annotation}
We decompose and annotate relevant object CAD models from AI2-THOR--Habitat dataset~\cite{khanna2023hssd} in Blender, separating them into semantically meaningful parts (\eg, cabinet body and handle, appliance body and buttons, knobs and burners). This produces a library of part-aware assets that can be instantiated in AI2-THOR scenes.

\paragraph{Ground-truth object and part point clouds.}
Using these annotated assets, we construct ground-truth 3D point clouds for both objects and parts within AI2-THOR scenes. For each object instance, we uniformly sample points along objects' and their parts' mesh surface, resulting in an object- and part-centric point cloud and object-part hierarchy.

\paragraph{Functional relation annotation.}
We then annotate functional relations using a set of predefined annotation rules. Each rule is defined as a functional triplet in the format (first label, relation, second label). The rules are grouped into categories based on the matching strategies used, as shown below:
\begin{enumerate}
  \item Exact matching - We annotate edges between node pairs with relation type ``relation'' if and only if the first node's label exactly matches the triplet's ``first label'' and the second node's label exactly matches the triplet's ``second label''. Representative examples of rules employing this strategy include: knife slices apple; faucet fills kettle with water.
  \item Proximity-based matching - For every node whose label matches the first label of the triplet, we identify the closest node matching the triplet's ``second label''. If such a node exists and the Euclidean distance between the node pair's centers is less than one meter, we annotate a functional relation labeled as ``relation'' between them. Some examples of rules using this strategy include: curtains cover/uncover windows; faucets fill sinks with water; and faucets fill bathtubs with water.
  \item Part-object and part-part matching - For objects that exhibit toggleable or openable properties in AI2-THOR and possess explicit functional part annotations such as power switches or handles (see Appendix \ref{app:part-annotation}), we annotate the corresponding node pairs with their respective functional relations. Representative rules in this category include: lever pushes down to activate toaster; handle pulls to open door.
  \item Manual matching - For semantically ambiguous relations, we manually record affected node pairs and store these annotations in a JSON file for subsequent loading by the annotation pipeline. Currently, only stove knob-burner associations require manual matching.
\end{enumerate}

\noindent These rules are applied consistently across scenes to produce dense and unbiased annotations for both part--object and object--object relations.

\paragraph{View sampling and RGB-D capture.}
To mimic realistic partial observations, we randomly sample reachable poses in each scene and a camera pitch angle for each pose. From these viewpoints, we capture RGB-D sequences and corresponding camera poses, with each scene containing 60 frames. These sequences form the input to our mapping and functional inference pipeline.

\paragraph{Visibility filtering.}
Not all objects are visible from the sampled camera trajectories (\eg, items stored inside closed cabinets). To ensure that evaluation only considers observable entities, we first fuse the posed RGB-D sequence into a class-agnostic point cloud and then retain only those ground-truth object and part points that lie within a fixed radius of this fused cloud. Objects whose visible subset contains fewer than 10 points are discarded, and any functional relations involving discarded objects are removed. We store the remaining visible nodes and their functional edges as the final \funthor benchmark used in our experiments.

\section{Instance Association and Merging Details}
\label{app:instance_association}
We adopt a two-stage instance association and merging strategy to consolidate object and part detections across multiple frames into a unified 3D object-part map. In the first stage, we perform spatial association by computing the Intersection over Union (IoU) between the 3D bounding boxes of newly detected instances and those of existing scene instances. If the IoU exceeds a threshold of $0.03$, the detections are considered to potentially correspond to the same instance, and we proceed to the second stage of semantic verification. In this stage, we use DINOv2 with registers \cite{darcet2023vitneedreg} to extract 2D features from the RGB image for each detection and compute the cosine similarity between the feature of the newly detected instance and the existing scene instance. If the similarity exceeds a threshold of $0.5$, we confirm that the two instances correspond to the same object or part and merge them by aggregating their point clouds. The label of the merged instance is then updated to reflect the most frequently occurring label among all detections composing the instance. If either the spatial association or semantic verification fails, the newly detected instance is treated as a new object or part and added to the scene map. It is worth noting that we only permit merging between instances of the same category (\ie, object-to-object or part-to-part) to prevent erroneous associations. Objects are merged prior to parts. When objects are merged, their contained parts are also aggregated and become parts of the merged object.

\section{Inference Time Analysis}
We analyze the inference time of \funfact on the FunGraph3D dataset, which consists of 14 scenes with an average of 203.3 RGB-D frames per scene. In total, \funfact processes all 14 scenes in 27479s ($\simeq7.6$h).

\paragraph{Scene reconstruction.} \funfact processes each frame in 9.7s on average, encompassing GPT-based scene analysis, object and part detection, instance association, and merging. Although \funfact queries GPT once per frame for scene analysis, this step accounts for nearly 93\% of the total stage runtime. Since scene analysis is performed independently for each frame, it can be straightforwardly parallelized across multiple concurrent GPT calls, which would significantly reduce the overall runtime.

\paragraph{Functional scene graph creation.} \funfact processes each scene in 26.3s on average, encompassing LLM-based functional proposal generation, edge and dual graph construction, and factor graph inference.

\section{Pseudo-code for Functional Scene Graph Construction}

A high-level pseudo-code for building the functional scene graph using \funfact is provided in Algorithm~\ref{alg:funfact}. The process consists of three main phases: (1) generating part-object and part-part connection factor graph (Lines~\ref{lst:phase1_start}--\ref{lst:phase1_end}), (2) generating object-object connection factor graph (Lines~\ref{lst:phase2_start}--\ref{lst:phase2_end}), and (3) performing global inference to estimate confidence scores and construct the final functional scene graph (Lines~\ref{lst:phase3_start}--\ref{lst:phase3_end}). At the beginning, we initialize two empty lists, $\mathcal{G}_L$ (Line~\ref{lst:local_group_list}) and $\mathcal{G}_R$ (Line~\ref{lst:remote_group_list}), to store EdgeGroups for part-object/part-part connections and object-object connections, respectively. For each LLM-proposed semantic functional proposal (\eg, remote control operates TV), we create an EdgeGroup that encapsulates all resulting edges (\eg, the edges connecting all pairs of remote controls and TVs) and their associated factor graph. The default edge confidence is set to the semantic confidence of the functional proposal. In phase 1, we iterate over each object with parts (Line~\ref{lst:part_loop}) and obtain LLM-proposed part-object and part-part connections (Line~\ref{lst:llm_propose_parts}). We assume proximity always holds for part-object and part-part connections, and we add a CardinalityFactor only when the proposal is believed by LLMs to be one-to-one (Line~\ref{lst:local_one_to_one}) (\ie the functional connection is mutually exclusive, such as a stove knob only controlling one stove burner). In phase 2, we first obtain LLM-proposed object-object connections (Line~\ref{lst:llm_propose_objects}) and build different factor graphs based on the nature of the proposed connections. If the proposal is one-to-one (Line~\ref{lst:remote_one_to_one}), we include both ProximityFactor and CardinalityFactor; if the proposal requires proximity but is not one-to-one (Line~\ref{lst:requires_proximity}), we only include ProximityFactor. Finally, in phase 3, we combine all EdgeGroups (Line~\ref{lst:combine_groups}) and their factor graphs (Line~\ref{lst:combine_factors}), and perform marginal inference (Line~\ref{lst:marginal_inference}) to estimate confidence scores for all edges. These scores are then used to update the confidence of each EdgeGroup (Line~\ref{lst:update_confidence}) before constructing the final functional scene graph (Line~\ref{lst:build_fsg}).

\begin{algorithm}[h!]
    \caption{Build Functional Scene Graph}\label{alg:funfact}
    \small
    \begin{algorithmic}[1]

        \Require Object map $\mathcal{O}$ with contained parts info
        \Ensure Functional scene graph

        \Statex
        \State $\mathcal{G}_L \gets []$\label{lst:local_group_list} \Comment{Local (part-object) based groups}
        \State $\mathcal{G}_R \gets []$\label{lst:remote_group_list}\Comment{Remote (object-object) groups}

        \Statex
        \State $\triangleright$ \textbf{Phase 1: Part-Object and Part-Part Connections}\label{lst:phase1_start}
        \For{each object $o \in \mathcal{O}$ with parts}\label{lst:part_loop}
        \State $\mathcal{P} \gets$ LLM-Propose-Parts($o$, $o$.parts)\label{lst:llm_propose_parts}
        \For{each proposal $p \in \mathcal{P}$}
        \State Create EdgeGroup $g$ from $p$
        \State Build FactorGraph $\mathcal{F}_g$ with:
        \State \quad \ ProximityFactor($g$) \Comment{Encodes spatial closeness}
        \If{$p$.isOneToOne}\label{lst:local_one_to_one}
        \State CardinalityFactor($g$)  \Comment{Enforce one-to-one}
        \EndIf
        \State $g.\mathcal{F} \gets \mathcal{F}_g$ \Comment{Attach factors to $g$}
        \State Add $g$ to $\mathcal{G}_L$ \Comment{Store local group}
        \EndFor
        \EndFor\label{lst:phase1_end}

        \Statex
        \State $\triangleright$ \textbf{Phase 2: Object-Object Connections}\label{lst:phase2_start}
        \State $\mathcal{R} \gets$ LLM-Propose-Objects($\mathcal{O}$)\label{lst:llm_propose_objects}\Comment{LLM object connection proposals}
        \For{each proposal $r \in \mathcal{R}$}
        \State Create EdgeGroup $g$ from $r$
        \If{$r$.isOneToOne}\label{lst:remote_one_to_one}
        \State Build FactorGraph $\mathcal{F}_g$ with:
        \State \quad ProximityFactor($g$)  \Comment{Encodes spatial closeness}
        \State \quad CardinalityFactor($g$)  \Comment{Enforce one-to-one}
        \State $g.\mathcal{F} \gets \mathcal{F}_g$
        \ElsIf{$r$.requiresProximity}\label{lst:requires_proximity}
        \State Build FactorGraph $\mathcal{F}_g$ with:
        \State \quad ProximityFactor($g$) \Comment{Encodes spatial closeness}
        \State $g.\mathcal{F} \gets \mathcal{F}_g$
        \EndIf
        \State Add $g$ to $\mathcal{G}_R$ \Comment{Store remote group}
        \EndFor\label{lst:phase2_end}

        \Statex
        \State $\triangleright\ $\textbf{ Phase 3: Global Inference}\label{lst:phase3_start}
        \State $\mathcal{G} \gets \mathcal{G}_L \cup \mathcal{G}_R$\label{lst:combine_groups}
        \State $\mathcal{F} \gets \bigcup_{g \in \mathcal{G}} g.\mathcal{F}$\label{lst:combine_factors}  \Comment{Global factor graph}
        \State $\mu \gets$ MarginalInference($\mathcal{F}$)\label{lst:marginal_inference}
        \State UpdateConfidence($\mathcal{G}$, $\mu$)\label{lst:update_confidence}
        \State $\mathcal{S} \gets$ BuildFSG($\mathcal{O}$,$\mathcal{G}$)\label{lst:build_fsg}\Comment{Final Functional Scene Graph}

        \State \Return $\mathcal{S}$\label{lst:phase3_end}

    \end{algorithmic}
\end{algorithm}

\section{Alternative VLM Backbones}
We evaluate the sensitivity of \funfact to the choice of VLM backbone by replacing GPT-4.1 with gemini-3-flash-preview and the open-weight qwen3-vl-32b-instruct. Scene reconstruction and triplet evaluation results are reported in Tab.~\ref{tab:reconstruction_vlm_suppl} and Tab.~\ref{tab:triplet_vlm_suppl}, respectively. \funfact performs consistently well across all three backbones, demonstrating that its strong performance stems from the pipeline design rather than reliance on any particular VLM.

\begin{table*}[ht!]
\centering
\caption{\textbf{Scene Reconstruction with Alternative VLM Backbones.} We report Recall@K (\textbf{R@3}/\textbf{R@10}; higher is better) for three node categories---\emph{Objects}, \emph{Interactive Elements}, and \emph{Overall Nodes}---on \textbf{SceneFun3D} and \textbf{FunGraph3D}, using gemini-3-flash-preview and qwen3-vl-32b-instruct as drop-in replacements for GPT-4.1. Bold numbers mark the best result per column.}
\renewcommand{\arraystretch}{1.1}
\resizebox{\textwidth}{!}{
\begin{tabular}{lcccccc|cccccc}
\toprule
\textbf{VLM Backbone} &
\multicolumn{6}{c|}{\textbf{SceneFun3D}} &
\multicolumn{6}{c}{\textbf{FunGraph3D}} \\
\cmidrule(lr){2-7}\cmidrule(lr){8-13}
& \multicolumn{2}{c}{Objects $(\uparrow)$}
& \multicolumn{2}{c}{Inter.\ Elements $(\uparrow)$}
& \multicolumn{2}{c|}{Overall Nodes $(\uparrow)$}
& \multicolumn{2}{c}{Objects $(\uparrow)$}
& \multicolumn{2}{c}{Inter.\ Elements $(\uparrow)$}
& \multicolumn{2}{c}{Overall Nodes $(\uparrow)$} \\
& R@3 & R@10 & R@3 & R@10 & R@3 & R@10
& R@3 & R@10 & R@3 & R@10 & R@3 & R@10 \\
\midrule
gemini-3-flash-preview  & \textbf{94.3} & 97.1 & 63.2 & \textbf{74.5} & \textbf{73.5} & \textbf{82.0} & \textbf{93.2} & \textbf{95.9} & \textbf{68.8} & \textbf{78.8} & \textbf{79.0} & \textbf{85.9} \\
qwen3-vl-32b-instruct   & 92.4 & \textbf{98.1} & \textbf{64.2} & 72.2 & \textbf{73.5} & 80.8 & 88.4 & 94.5 & 59.4 & 73.3 & 71.6 & 82.2 \\
\bottomrule
\end{tabular}
}
\label{tab:reconstruction_vlm_suppl}
\end{table*}

\begin{table*}[ht!]
\centering
\caption{\textbf{Triplet Evaluation with Alternative VLM Backbones.} We report node association, edge prediction and overall triplet recall as Recall@K (R@5/R@10) on \textbf{SceneFun3D} and \textbf{FunGraph3D}, and (R@3/R@5) on \textbf{FunThor}, using gemini-3-flash-preview and qwen3-vl-32b-instruct as drop-in replacements for GPT-4.1. Bold numbers mark the best result per column.}
\renewcommand{\arraystretch}{1.1}
\resizebox{\textwidth}{!}{
\begin{tabular}{lcccccc|cccccc|cccccc}
\toprule
\textbf{VLM Backbone} &
\multicolumn{6}{c|}{\textbf{SceneFun3D}} &
\multicolumn{6}{c|}{\textbf{FunGraph3D}} &
\multicolumn{6}{c}{\textbf{FunThor}} \\
\cmidrule(lr){2-7}\cmidrule(lr){8-13}\cmidrule(l){14-19}
& \multicolumn{2}{c}{Node Assoc.\ ($\uparrow$)} & \multicolumn{2}{c}{Edge Pred.\ ($\uparrow$)} & \multicolumn{2}{c|}{Overall Triplets ($\uparrow$)}
& \multicolumn{2}{c}{Node Assoc.\ ($\uparrow$)} & \multicolumn{2}{c}{Edge Pred.\ ($\uparrow$)} & \multicolumn{2}{c|}{Overall Triplets ($\uparrow$)}
& \multicolumn{2}{c}{Node Assoc.\ ($\uparrow$)} & \multicolumn{2}{c}{Edge Pred.\ ($\uparrow$)} & \multicolumn{2}{c}{Overall Triplets ($\uparrow$)} \\
& R@5 & R@10 & R@5 & R@10 & R@5 & R@10
& R@5 & R@10 & R@5 & R@10 & R@5 & R@10
& R@3 & R@5  & R@3 & R@5  & R@3 & R@5 \\
\midrule
gemini-3-flash-preview &
\textbf{67.2} & \textbf{72.3} & \textbf{51.1} & \textbf{68.1} & \textbf{34.4} & \textbf{49.2} &
\textbf{74.3} & \textbf{79.6} & \textbf{61.5} & 72.1 & \textbf{45.2} & \textbf{57.4} &
\textbf{70.9} & \textbf{72.3} & \textbf{76.2} & \textbf{78.5} & \textbf{54.1} & \textbf{56.8} \\
qwen3-vl-32b-instruct &
66.2 & 69.7 & 46.5 & 67.7 & 30.8 & 47.2 &
69.6 & 74.8 & 60.6 & \textbf{73.8} & 42.2 & 55.2 &
68.9 & 70.9 & 72.5 & 78.1 & 50.0 & 55.4 \\
\bottomrule
\end{tabular}
}
\label{tab:triplet_vlm_suppl}
\end{table*}

\section{Exclusive and Non-Exclusive Matching}
\begin{table*}[!t]
\centering
\caption{\textbf{Mapping Quality under Exclusive Matching.} We report Recall@K (\textbf{R@3}/\textbf{R@10}; higher is better) for three node categories—\emph{Objects}, \emph{Interactive Elements}, and \emph{Overall Nodes}—on \textbf{SceneFun3D}, \textbf{FunGraph3D}, and \textbf{FunThor} under the exclusive matching constraint, where each predicted node maps to at most one ground-truth node. This provides a more accurate assessment of mapping quality compared to non-exclusive matching, which may overestimate performance when a single merged detection covers multiple ground-truth objects. All methods exhibit decreased recall under this stricter evaluation, but \textbf{FunFact} maintains superior performance across nearly all categories and datasets.}
\renewcommand{\arraystretch}{1.1}
\resizebox{1.0\textwidth}{!}{
\begin{tabular}{lcccccc|cccccc|cccccc}
\toprule
\textbf{Methods} &
\multicolumn{6}{c|}{\textbf{SceneFun3D}} &
\multicolumn{6}{c|}{\textbf{FunGraph3D}} &
\multicolumn{6}{c}{\textbf{FunThor}} \\
\cmidrule(lr){2-7}\cmidrule(lr){8-13}\cmidrule(lr){14-19}
& \multicolumn{2}{c}{Objects $(\uparrow)$ } & \multicolumn{2}{c}{Inter.\ Elements $(\uparrow)$ } & \multicolumn{2}{c|}{Overall Nodes $(\uparrow)$ }
& \multicolumn{2}{c}{Objects $(\uparrow)$ } & \multicolumn{2}{c}{Inter.\ Elements $(\uparrow)$ } & \multicolumn{2}{c|}{Overall Nodes $(\uparrow)$ }
& \multicolumn{2}{c}{Objects $(\uparrow)$ } & \multicolumn{2}{c}{Inter.\ Elements $(\uparrow)$ } & \multicolumn{2}{c}{Overall Nodes $(\uparrow)$ }
 \\
& R@3 & R@10 & R@3 & R@10 & R@3 & R@10
& R@3 & R@10 & R@3 & R@10 & R@3 & R@10
& R@3 & R@10 & R@3 & R@10 & R@3 & R@10
 \\
\midrule
OpenFunGraph       & {78.1} & {79.0} & \textbf{64.4} & \textbf{78.3} & {69.1} & {78.5}
                   & {61.1} & {68.5} & {42.1} & {52.5} & {50.3} & {59.2}
                   & {43.7} & {52.3} & {34.8} & {40.4} & {41.5} & {49.3}
            \\
\midrule

 FunFact (Ours)    & \textbf{90.5} & \textbf{90.5} & \textbf{64.4} & {75.5} & \textbf{73.2} & \textbf{80.4}
                   & \textbf{84.9} & \textbf{89.0} & \textbf{62.9} & \textbf{75.2} & \textbf{72.1} & \textbf{81.0}
                   & \textbf{57.2} & \textbf{61.0} & \textbf{64.5} & \textbf{68.8} & \textbf{59.1} & \textbf{63.0} \\

\bottomrule
\end{tabular}
}
\label{tab:exclusive_reconstruction_result}
\end{table*}

When calculating recall to evaluate the mapping quality, following the evaluation protocol in \cite{zhang2025open}, it is necessary to count how many ground-truth nodes are correctly mapped to predicted nodes. However, \cite{zhang2025open}'s protocol allows multiple ground-truth nodes to map to the same predicted node, which may lead to an overestimation of mapping quality. For instance, when multiple cabinets are located in close proximity, a single merged detection that spatially covers all cabinets may achieve perfect recall, as each ground-truth cabinet can map to the same predicted node. To more accurately assess mapping quality, we additionally evaluate the results under an exclusive matching constraint, where each predicted node maps to at most one ground-truth node. Table \ref{tab:exclusive_reconstruction_result} presents the results under this setting. We observe a drop in recall for all methods under exclusive matching, indicating that some predicted nodes are mapped to multiple ground-truth nodes under non-exclusive matching. However, the relative performance between different methods remains consistent, with our method outperforming the baselines by a significant margin except Recall@10 for interactive elements in SceneFun3D.

\section{Confidence Histograms and Reliability Diagrams}

\begin{figure}[!t]
    \centering

    \begin{minipage}[t]{0.23\textwidth}
        \centering
        \includegraphics[width=\linewidth]{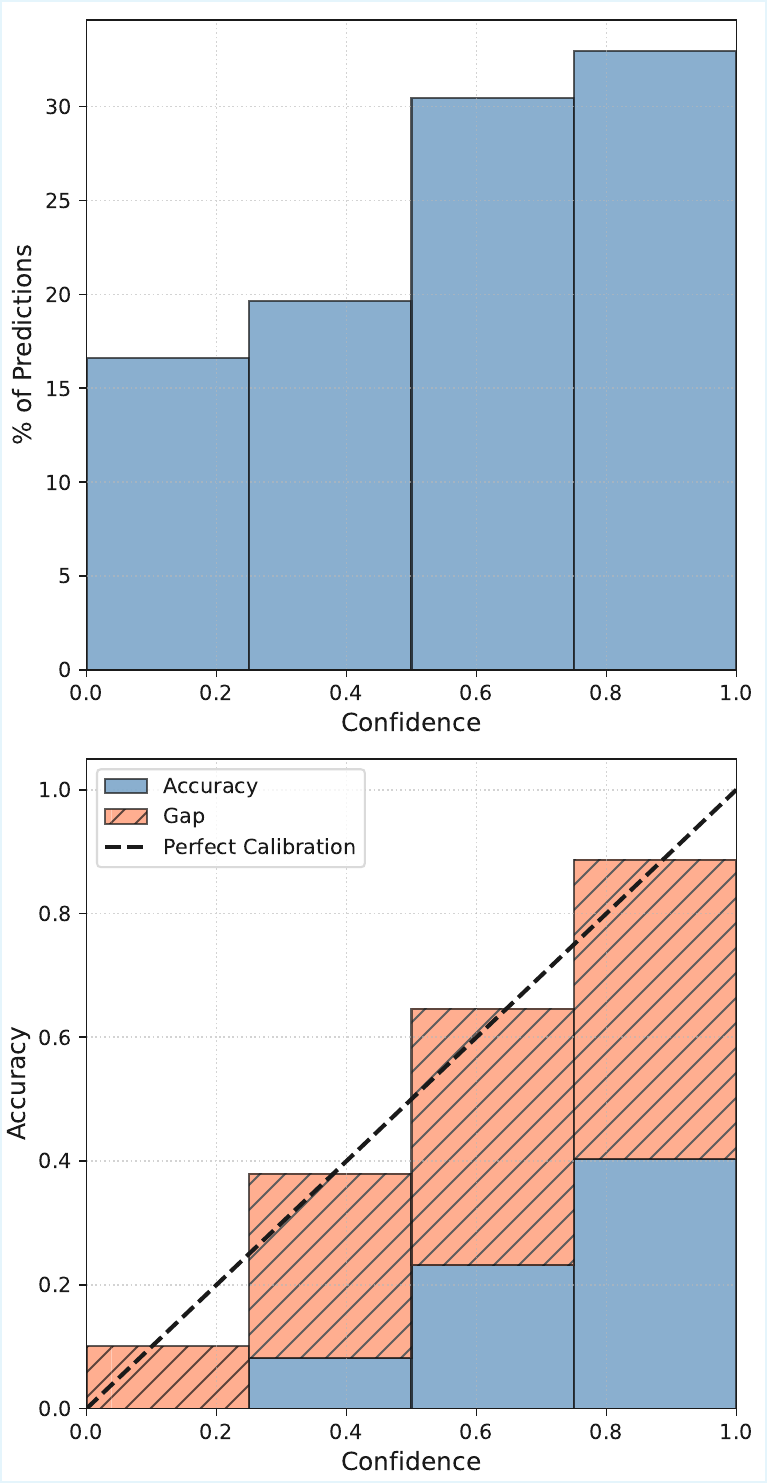}
        \subcaption{All predictions (Ours)}
    \end{minipage}
    \hfill
    \begin{minipage}[t]{0.23\textwidth}
        \centering
        \includegraphics[width=\linewidth]{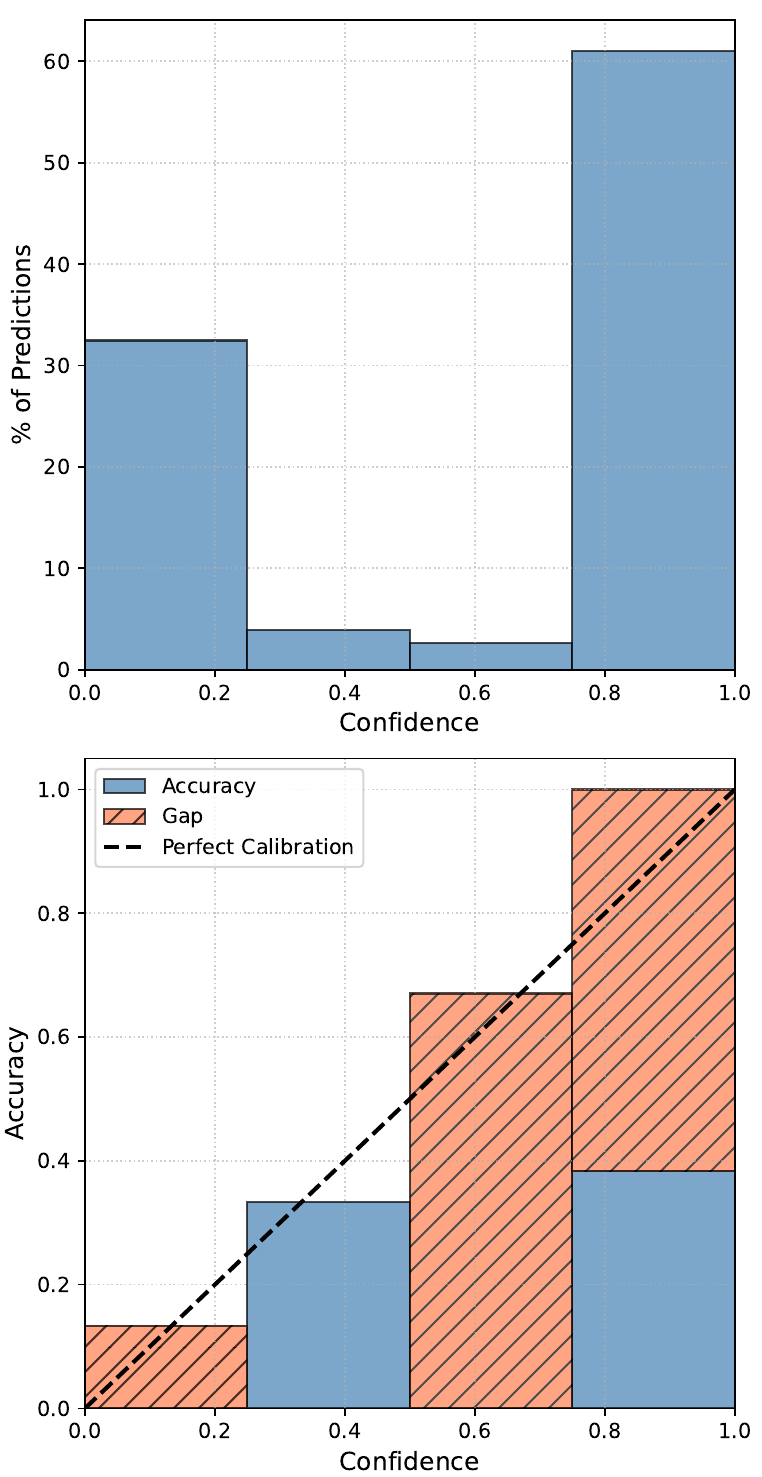}
        \subcaption{All predictions (OpenFunGraph)}
    \end{minipage}

    \vspace{3cm}

    \begin{minipage}[t]{0.23\textwidth}
        \centering
        \includegraphics[width=\linewidth]{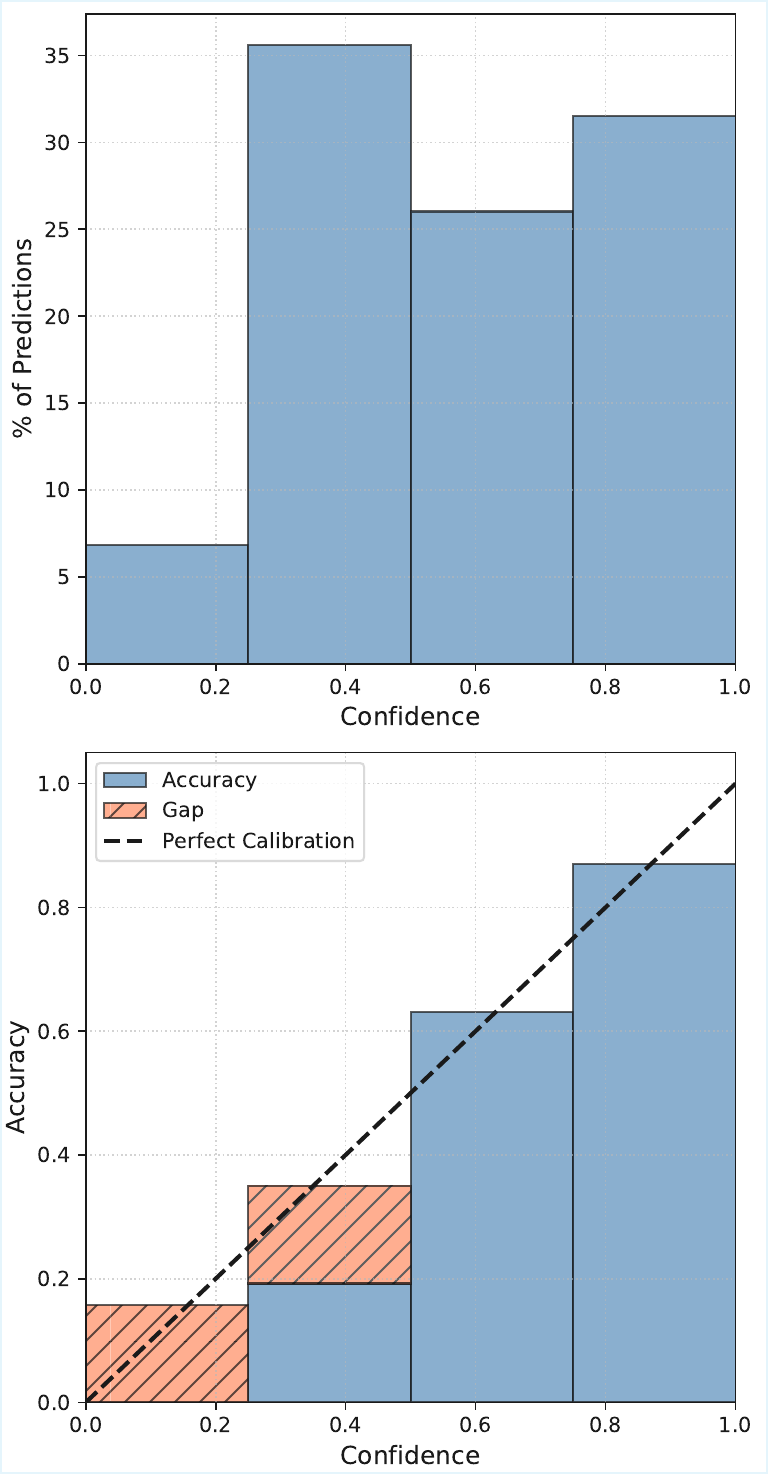}
        \subcaption{Ambiguous classes only (Ours)}
    \end{minipage}
    \hfill
    \begin{minipage}[t]{0.23\textwidth}
        \centering
        \includegraphics[width=\linewidth]{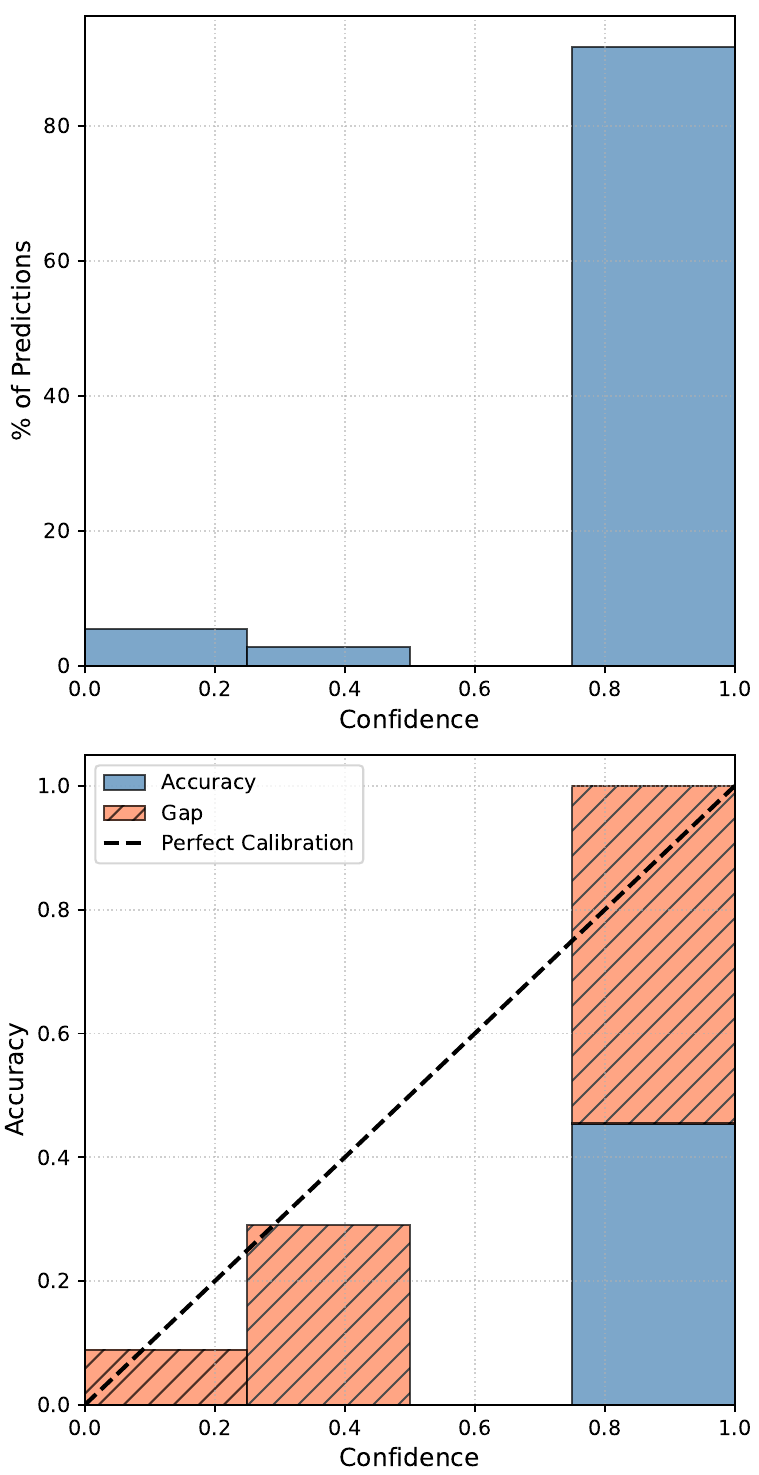}
        \subcaption{Ambiguous classes only (OpenFunGraph)}
    \end{minipage}

    \caption{\textbf{Calibration analysis on \funthor dataset.} Each subfigure shows a confidence histogram (top) and reliability diagram (bottom).}\label{fig:confidence_reliability_diagram}
\end{figure}

Fig.~\ref{fig:confidence_reliability_diagram} presents confidence histograms and reliability diagrams for \funfact and OpenFunGraph on the \funthor dataset, providing a visual assessment of model calibration \cite{niculescu2005predicting}. The confidence histogram illustrates the distribution of predicted confidence scores, while the reliability diagram compares predicted confidence against actual accuracy. We follow the procedure in \cite{guo2017calibration} to plot the figures. We use a bin size of 4 (\ie, each bin spans a confidence interval of 0.25) to be consistent with the settings in Section~\ref{subs:aithor}. In the reliability diagram, the diagonal line represents perfect calibration, and the red bars indicate the gap between average confidence and accuracy within each bin. Since OpenFunGraph estimates confidence scores only for classes containing ``outlet,'' ``switch,'' ``power,'' and ``remote,'' we assign a confidence of 1.0 to all other predictions when constructing these diagrams.

When considering all predictions, \funfact produces a more uniform distribution of confidence scores across the full range, whereas OpenFunGraph assigns confidence scores only to a fixed subset of classes. Both models exhibit overconfidence and achieve comparable performance in the high-confidence range. This behavior is expected, as both methods rely primarily on visual data and LLM priors for functional relation prediction, with limited visual cues available to assess the reliability of high-confidence predictions.

However, when examining ambiguous classes specifically, \funfact demonstrates superior calibration compared to OpenFunGraph, yielding confidence scores that better align with actual accuracy. This improvement stems from \funfact's use of a factor graph to estimate confidence scores holistically across all edges, incorporating global context and inter-dependencies among multiple functional relations. For instance, when multiple switches and lights coexist in a scene, \funfact can reason about mutual exclusivity among different switch-light pairs, resulting in better-calibrated confidence scores. In contrast, OpenFunGraph makes independent predictions for each object pair without considering broader contextual information, leading to less reliable confidence estimates for ambiguous classes.

\section{Additional Qualitative Results}
\label{sec:qualitative_results}

\begin{figure*}[!t]
\centering
\includegraphics[width=0.9\linewidth]{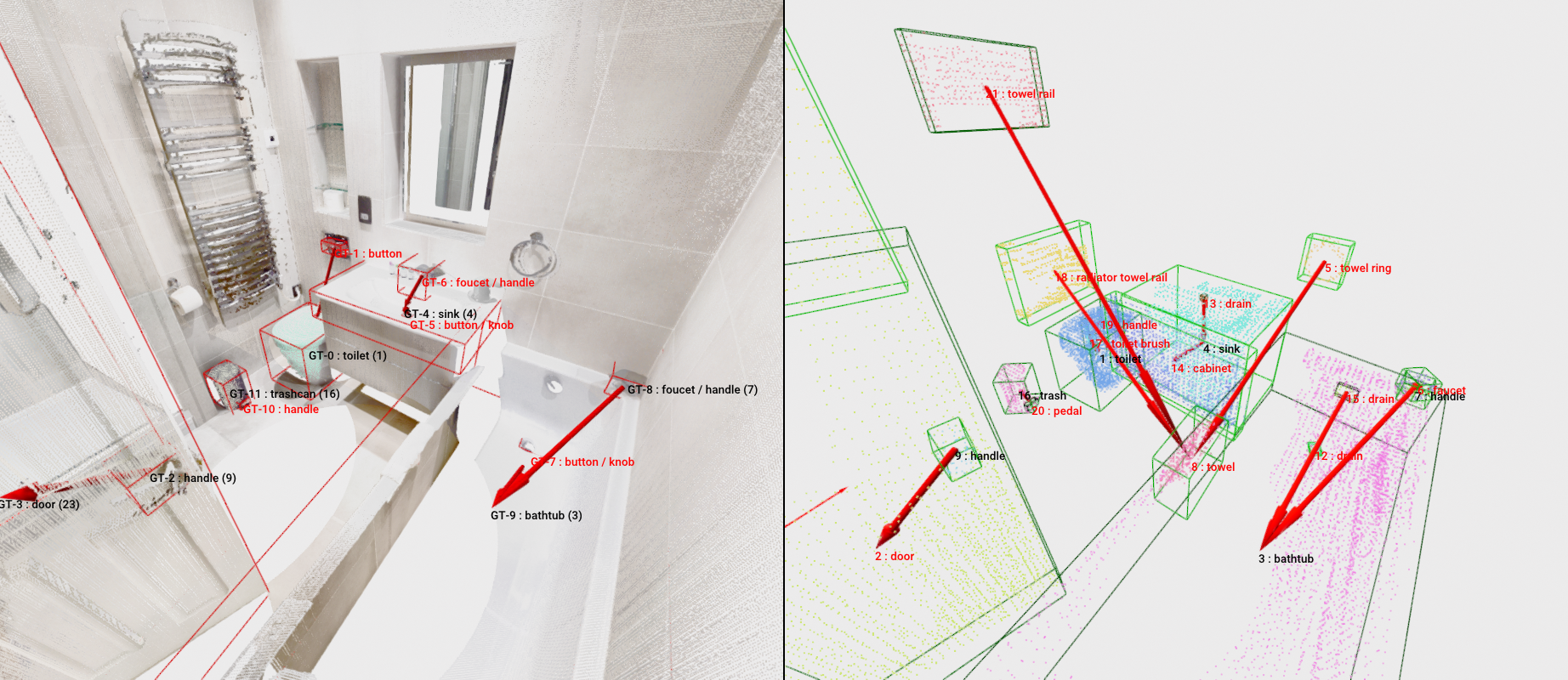}
\vspace{0.5cm}
\includegraphics[width=0.9\linewidth]{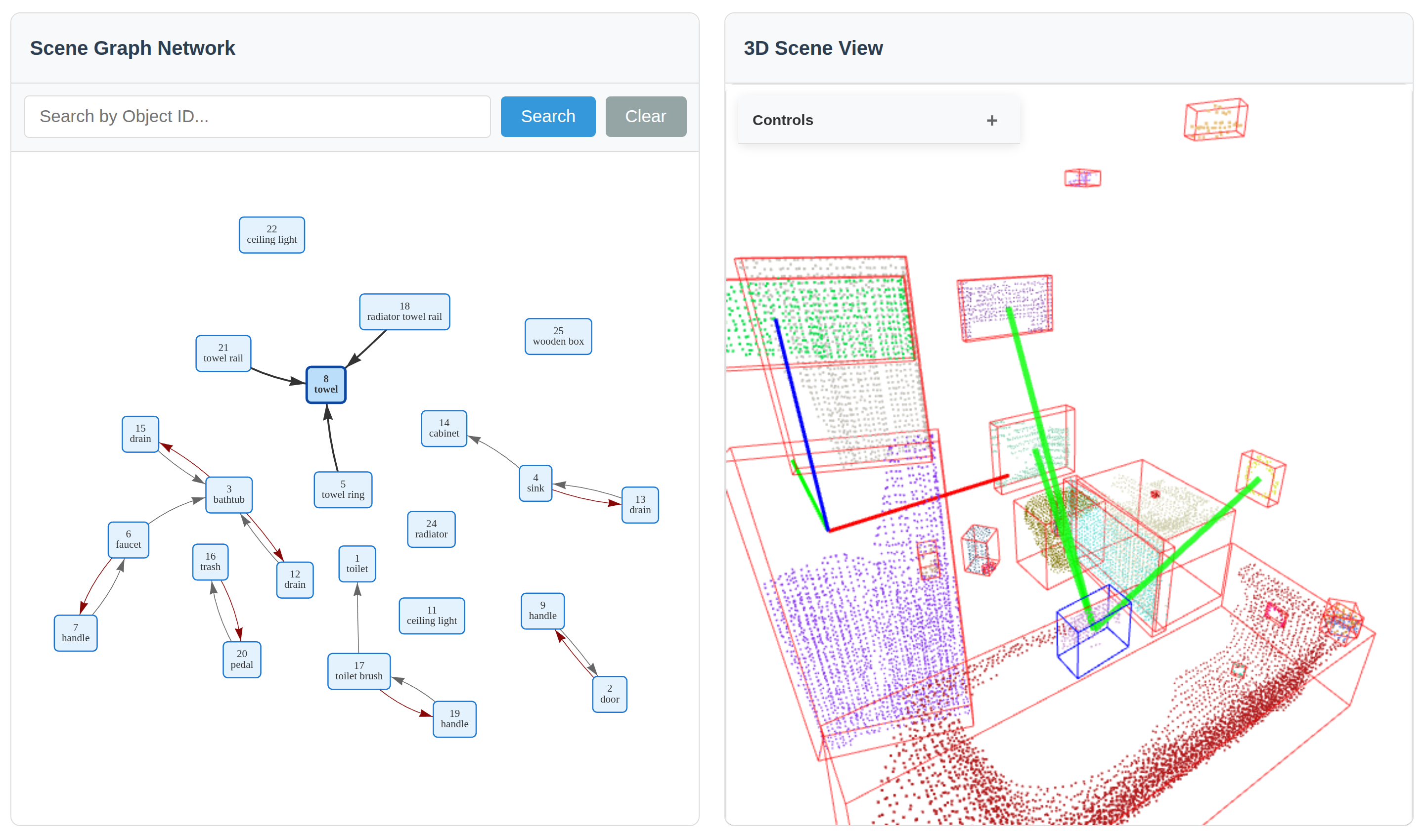}
\caption{\textbf{Scene \textit{dev/421063/42444511} of SceneFun3D.} See Section~\ref{sec:qualitative_results} for detailed subfigure descriptions. As shown in the top-right subfigure, we predict semantically correct functional relations in this bathroom scene, but none of them are matched to ground truth as indicated by the red arrows. This is due to several annotation issues in the dataset: (1) node label mismatches where ``drains'' are incorrectly labeled as ``button/knob'' and the trash bin ``pedal'' is labeled as ``handle'' in ground truth; (2) the ground truth label ``foucet'' is a typo and can never match our correct prediction ``faucet''; (3) even when node labels match, our open-vocabulary relation prediction ``handle turn or pull to open or close door'' cannot be matched to the ground truth annotation ``handle rotate to open or close door'' because when computing semantic similarity using BERT embeddings, the ground truth relation does not rank in the top-5 most similar relations, despite clearly describing the same interaction to humans. These issues lead to zero matched relations despite semantic correctness. Beyond ground truth, our method identifies additional functional relations, including ``towel ring can hold towel'' and ``towel radiator can dry towel'' (highlighted in bottom-left graph and bottom-right point cloud viewer). \vspace{1.5cm}
}
\label{fig:qual_scenefun3d_1}
\end{figure*}

\begin{figure*}[!t]
\centering
\includegraphics[width=0.9\linewidth]{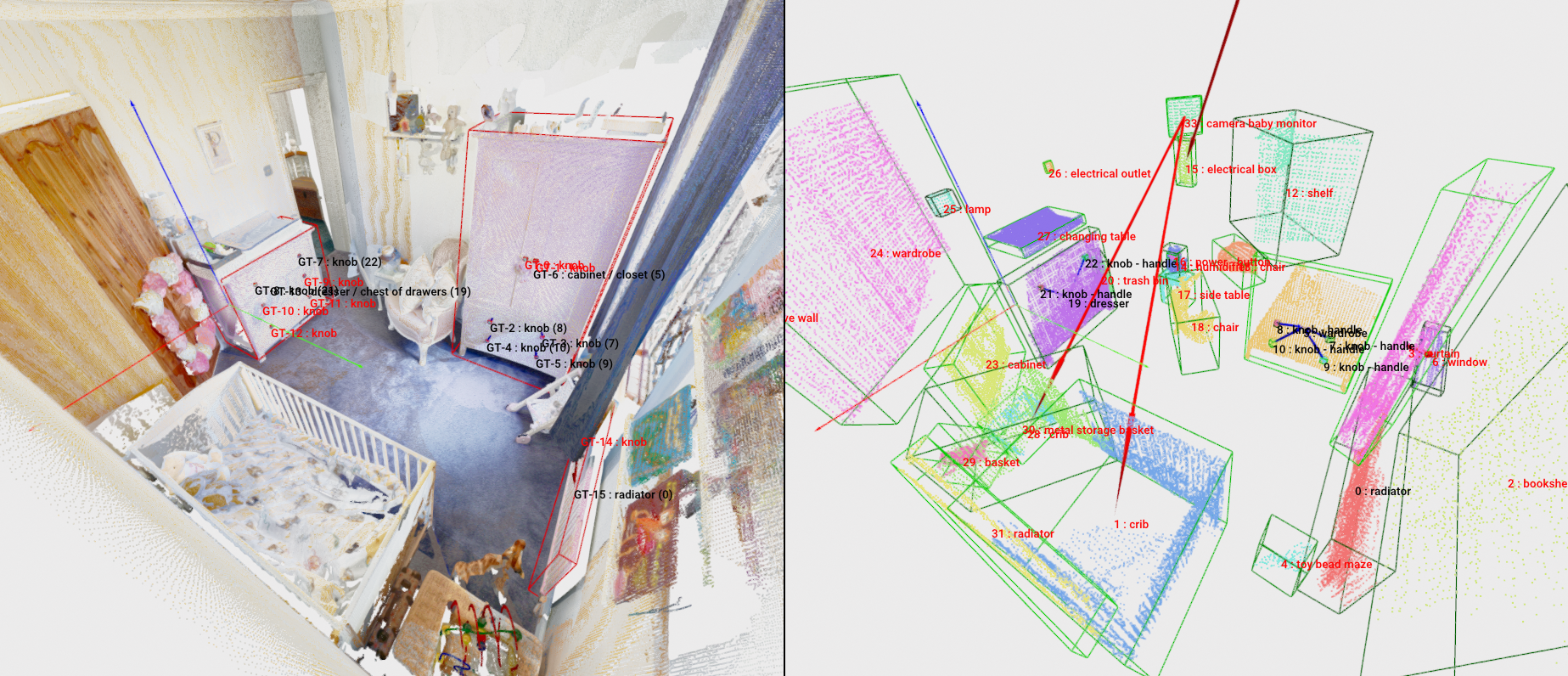}
\vspace{0.5cm}
\includegraphics[width=0.9\linewidth]{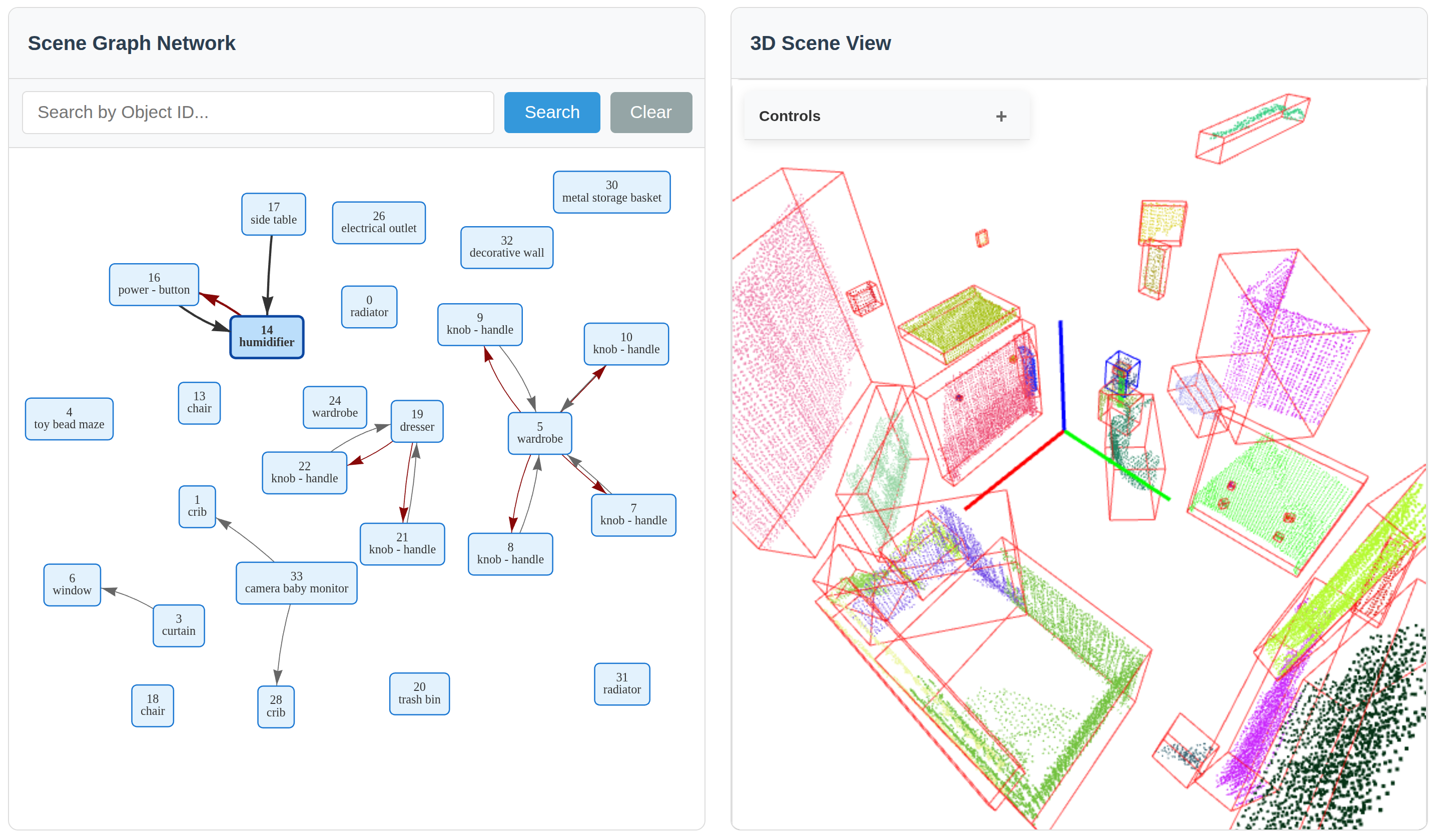}
\caption{\textbf{Scene \textit{dev/422007/42446017} of SceneFun3D.} See Section~\ref{sec:qualitative_results} for detailed subfigure descriptions. This baby bedroom scene highlights how the SceneFun3D dataset overly focuses on knob-pull interactions to open cabinets and drawers. Nearly all ground-truth annotations involve such interactions, with only one exception for radiator temperature adjustment. The dataset overlooks inter-object and intra-object functional relations that are crucial for understanding the scene. As shown in the top-right subfigure (predictions) and bottom-left subfigure (functional scene graph), our method identifies these missing relations, including ``power button press to turn on/off humidifier'' and ``baby monitor camera watch crib'', demonstrating our pipeline's ability to capture diverse functional interactions beyond the narrow focus of existing annotations.\vspace{2cm}}
\label{fig:qual_scenefun3d_2}
\end{figure*}

\begin{figure*}[!t]
\centering
\includegraphics[width=0.9\linewidth]{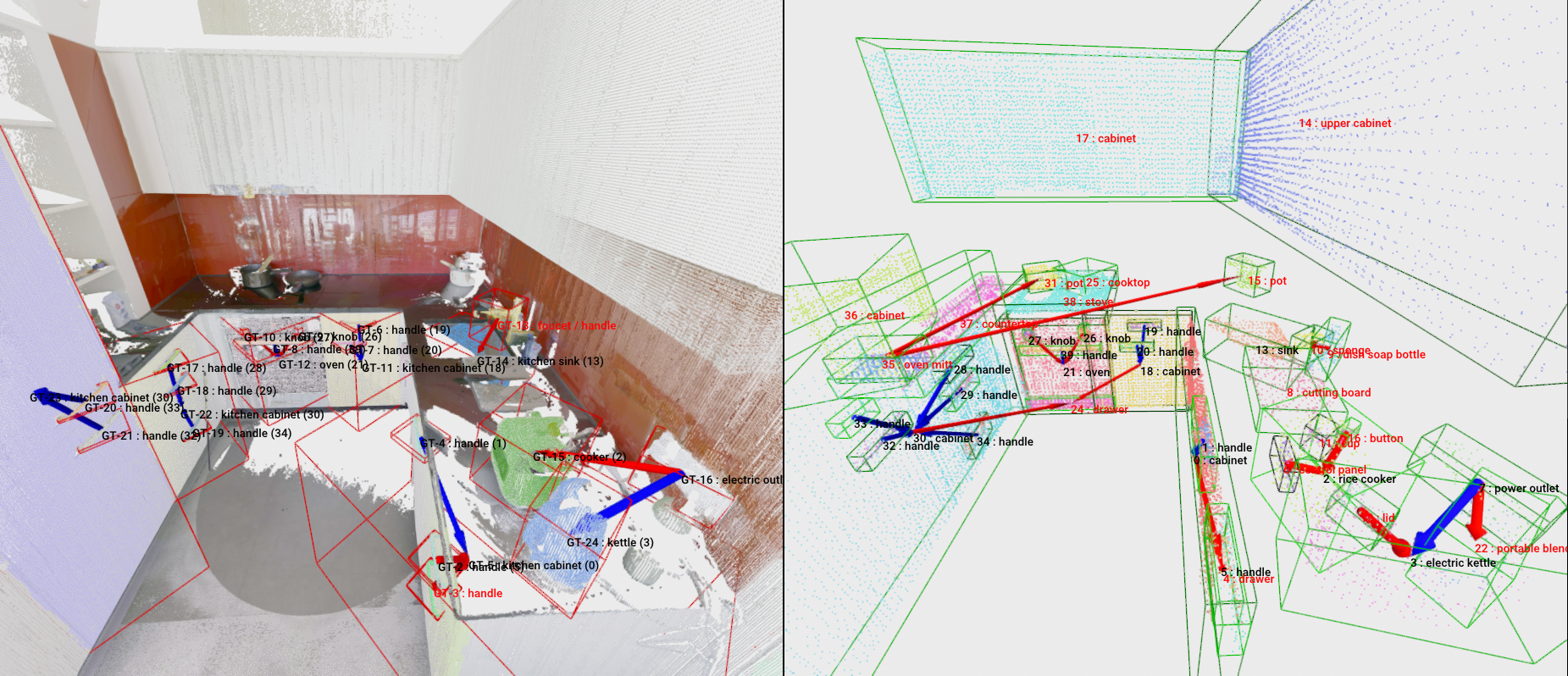}
\vspace{0.5cm}
\includegraphics[width=0.9\linewidth]{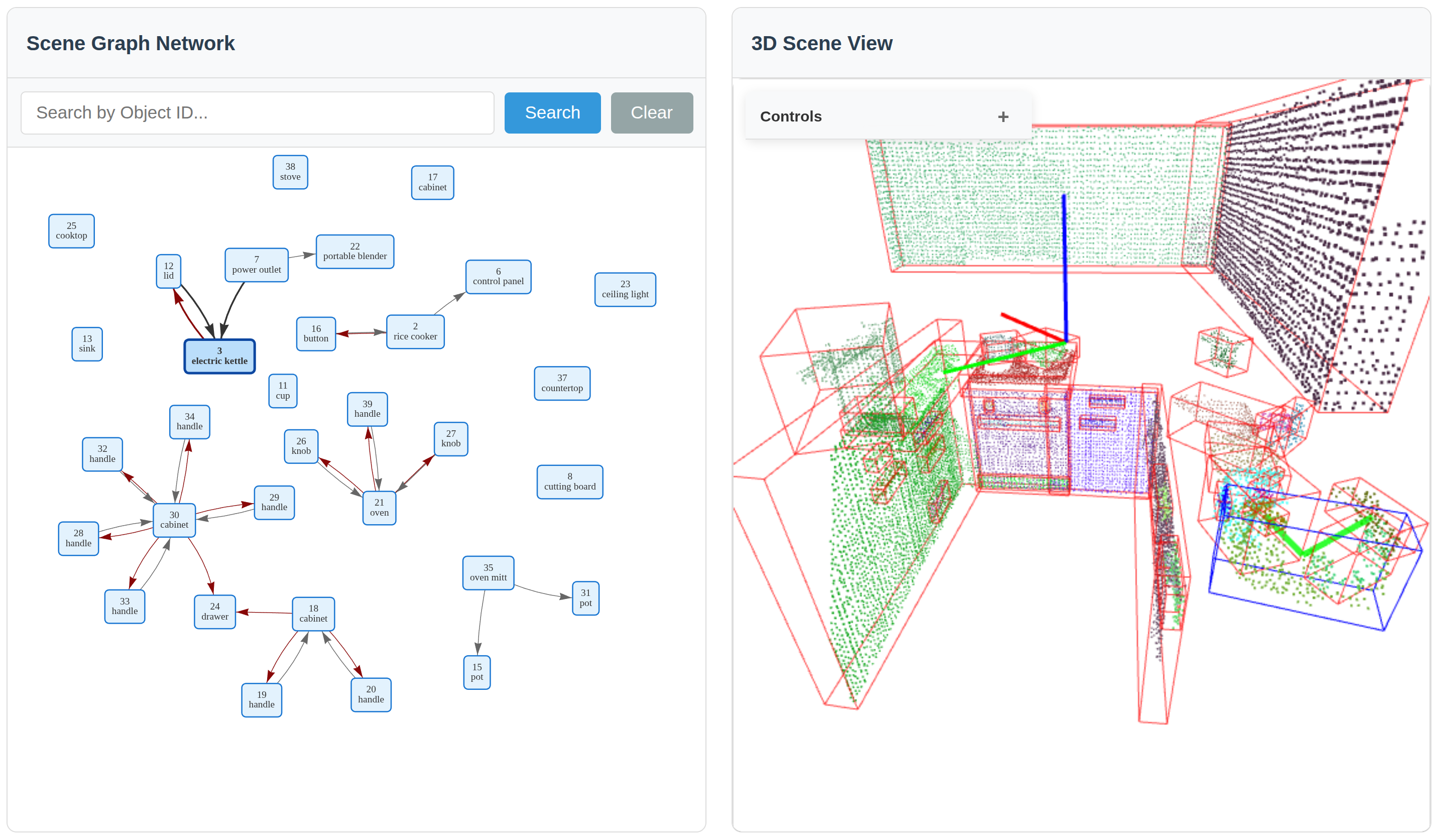}
\caption{\textbf{Scene \textit{4livingroom} of FunGraph3D.} See Section~\ref{sec:qualitative_results} for detailed subfigure descriptions. Our pipeline predicts most ground-truth functional relations, including power outlet-to-kettle connections and oven knob temperature/mode adjustments. We highlight ``kettle'' in the bottom two subfigures to demonstrate that our pipeline captures both inter-object relations (\eg, power outlet provides power to kettle) and intra-object relations (\eg, lid flip to open kettle). \vspace{4cm}
}
\label{fig:qual_fungraph}
\end{figure*}

\begin{figure*}[!t]
\centering
\includegraphics[width=0.9\linewidth]{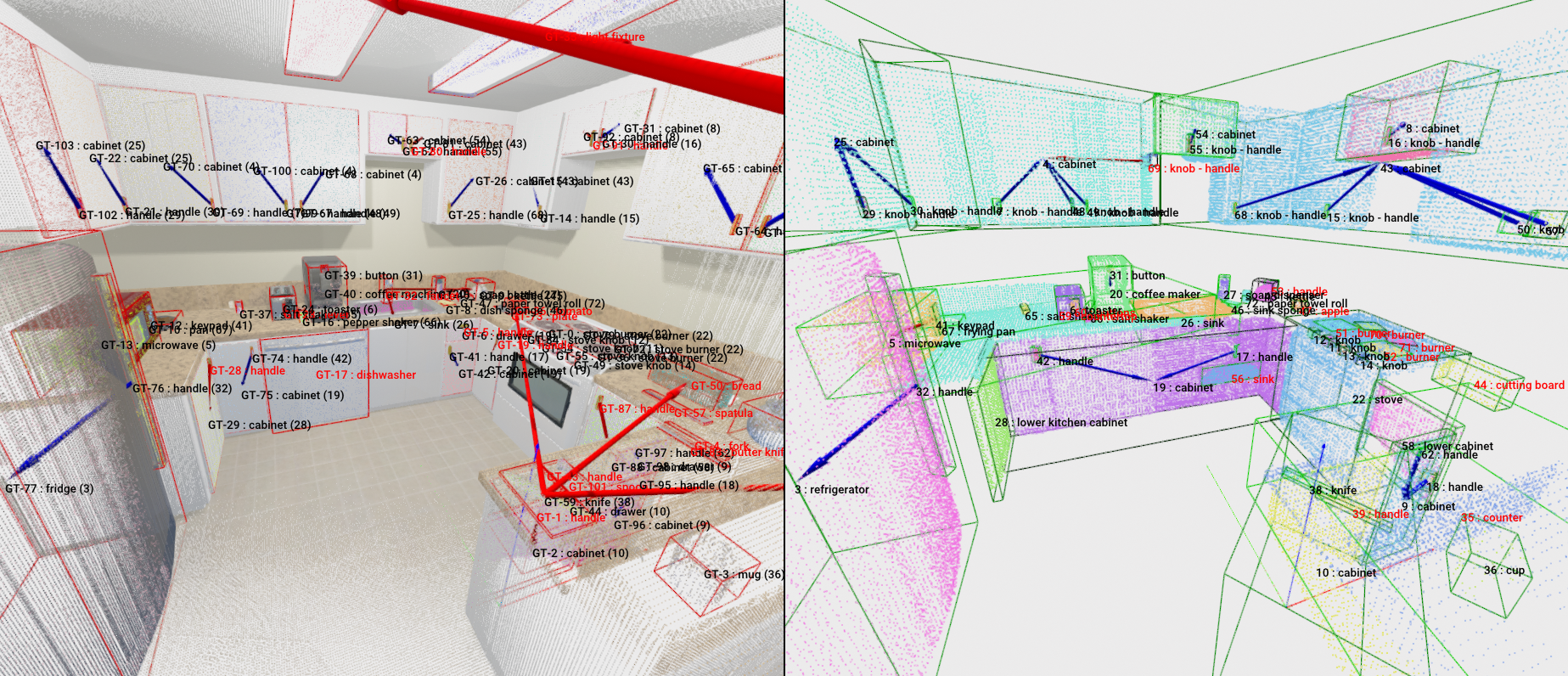}
\includegraphics[width=0.9\linewidth]{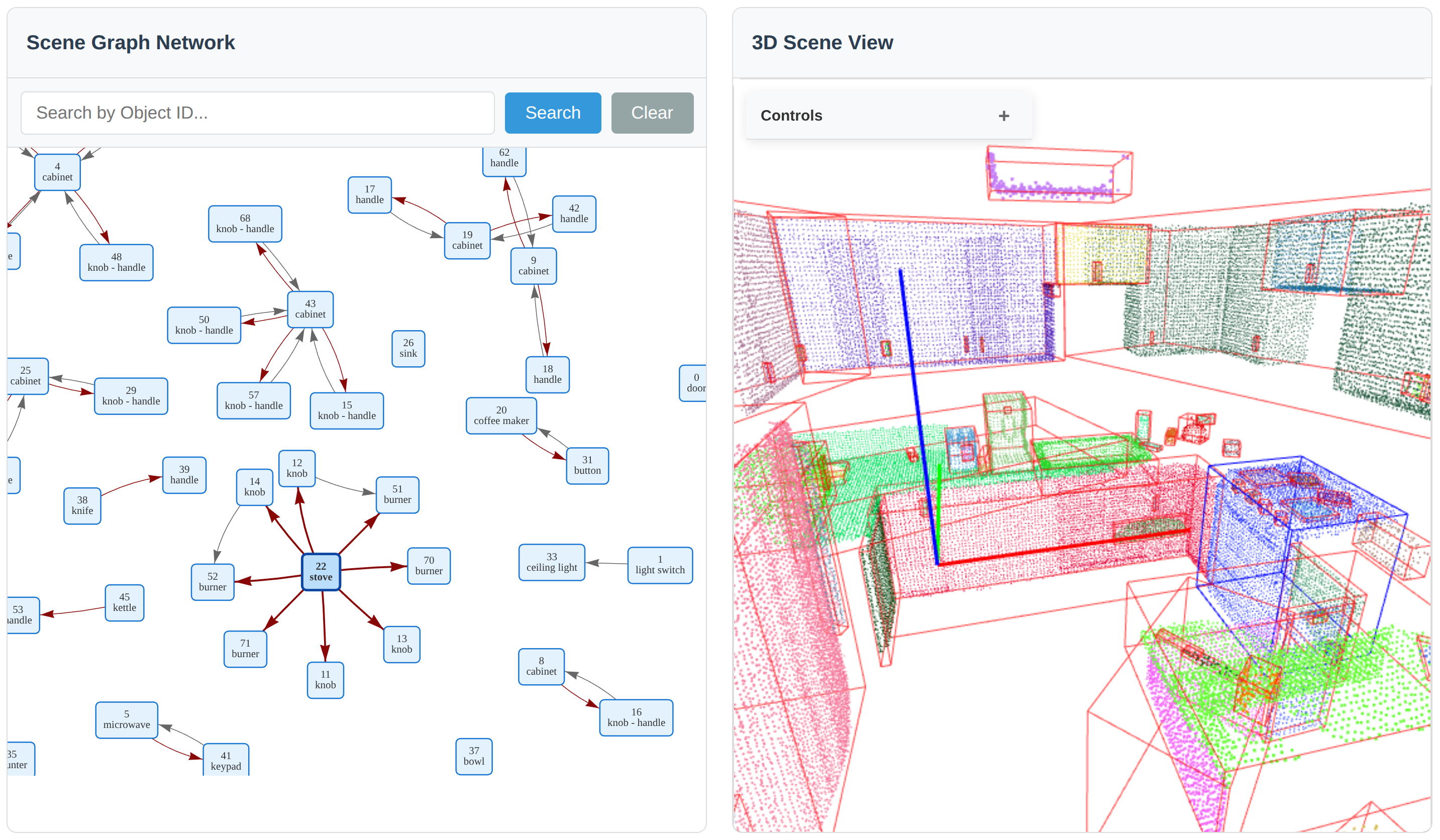}
\caption{\textbf{Scene \textit{FloorPlan5} of \funthor.} See Section~\ref{sec:qualitative_results} for detailed subfigure descriptions. We illustrate the dense and complete ground-truth annotations for the \funthor dataset (top-left). Our pipeline accurately detects functional parts of objects (\eg, burners and knobs of the stove; buttons of the coffee machine). As only predicted functional relations with confidence scores above $0.5$ are displayed in the functional scene graph, only the two outermost knobs are predicted to functionally connect to their respective outermost burners in the graph. \vspace{3cm}}

\label{fig:qual_funthor}
\end{figure*}

In this section, we present qualitative results for both the mapping stage and functional scene graph construction stage of \funfact on the SceneFun3D (Fig.~\ref{fig:qual_scenefun3d_1} and \ref{fig:qual_scenefun3d_2}), FunGraph3D (Fig.~\ref{fig:qual_fungraph}), and \funthor datasets (Fig.~\ref{fig:qual_funthor}). Each figure comprises four subfigures arranged from left to right and top to bottom as follows: \textbf{(a: top-left)} ground truth point cloud with node and functional relation annotations; \textbf{(b: top-right)} reconstructed object- and part-centric point cloud with predicted node labels and functional relations; \textbf{(c: bottom-left)} predicted functional scene graph with one node selected (highlighted); \textbf{(d: bottom-right)} visualization of the selected node in the reconstructed point cloud along with all functional relations associated with that node.

In subfigures (a) and (b), black node labels indicate matched nodes, while red node labels indicate unmatched nodes. Similarly, blue arrows denote matched functional relations, and red arrows denote unmatched functional relations. In subfigure (c), the selected node is highlighted in dark blue, red edges indicate object-part hierarchical relationships (always directed from object to part), and gray edges represent functional relations. In subfigure (d), the bounding box of the selected node is highlighted in blue, and all functional relations involving the selected node are visualized using green lines.

For the SceneFun3D and FunGraph3D datasets, our reconstruction quality is generally high, with most objects and parts accurately detected and labeled. Our pipeline consistently identifies more functional elements and functional relations than those annotated in the ground truth, revealing the sparse annotation limitations of existing datasets. In Fig.~\ref{fig:qual_scenefun3d_1}, we demonstrate some misalignment between predicted node labels and ground truth labels, where all drains are labeled as ``button/knob'' and the trash bin pedal is labeled as ``handle'' in the ground truth. Due to this misalignment, some functional relations are incorrectly marked as unmatched despite being correctly predicted. This issue highlights the limitations of existing evaluation protocols for open-vocabulary functional scene graphs.

For the \funthor dataset, our method effectively reconstructs objects and parts in complex indoor scenes, accurately capturing functional relations such as stove knobs operating burners and light switches controlling lights. These qualitative results demonstrate the effectiveness of \funfact in mapping and understanding functional scene graphs across diverse datasets.

\section{Potential Pipeline Adaptation for SceneFun3D dataset}
As discussed in Section~\ref{subsec:mapping}, SceneFun3D primarily focuses on part annotations, especially on knob and handle parts. It systematically annotates knob-shaped components (\eg, sink drains and radiator valves) with the generic label ``knob'', leading to lexical discrepancies between open-vocabulary predictions and ground-truth labels, thereby negatively impacting both mapping and functional relation evaluation. To isolate the effects of the evaluation protocol from actual pipeline performance on this issue (\ie, ``knob'' and ``drain'' are semantically far apart in CLIP embedding space), we modify the mapping pipeline to always detect knobs and handles whenever the scene analysis VLM identifies any functional parts in an object. This adaptation yields two key effects: first, it forces detection of knobs and handles independent of VLM proposals, thereby increasing recall for these part categories; second, it promotes assignment of generic ``knob'' and ``handle'' labels to merged knob- and handle-shaped components, as these labels become more prevalent during the merging process and are selected as the most common label (see Section~\ref{app:instance_association} for merging details).

Table~\ref{tab:fine_tune_on_scenefun3d} presents results on SceneFun3D after pipeline adaptation. Both mapping and functional relation metrics improve substantially; notably, the modified pipeline surpasses OpenFunGraph on all mapping-related metrics. Functional relation recall increases significantly due to improved part detection with aligned labels. However, overall triplet Recall@5 remains below OpenFunGraph, primarily due to semantic variations in relation descriptions. For instance, the ground-truth relation ``handle rotate to open or close door'' is functionally equivalent to our predicted ``handle turn or pull to open or close door,'' yet not ranked in the top-5 retrieval during evaluation. When relaxing the evaluation to consider top-10 retrieval, our method achieves an overall triplet recall of 75.9\%, outperforming OpenFunGraph's 70.3\%. This performance reversal highlights the sensitivity of triplet recall to top-k selection and suggests that our predictions capture correct semantics despite lexical differences.

However, enforcing knob and handle detection can introduce false positives, which indirectly affects functional relation precision. To assess the impact of this adaptation, we evaluate the modified pipeline on \funthor for both mapping and functional relation metrics, as shown in Table~\ref{tab:eval_fine_tuned}. The results indicate that the adaptation indeed significantly improves the recall of interactive elements and functional relations, but also leads to a substantial decrease in precision and F1 score for functional relations. Nevertheless, the modified pipeline still achieves better F1 score for functional relation than OpenFunGraph (25.1\% vs. 16.0\%) on \funthor. Given the reduced precision, however, we adopt the original pipeline for general applications, while the modified version can be beneficial for datasets with known annotation patterns.

\begin{table*}[!t]
\centering
\caption{\textbf{Performance after pipeline adaptation on SceneFun3D.} To address SceneFun3D's systematic annotation of knob-shaped and handle-shaped components (\eg, drains, valves, lever) with generic ``knob'' and ``handle'' labels, we modify our pipeline to always detect knobs and handles when functional parts are identified. The modified pipeline substantially improves both mapping and functional relation metrics, surpassing OpenFunGraph on all mapping metrics and achieving competitive triplet recall, especially at R@10 (75.9\% vs. 70.3\%).}
\resizebox{0.95\textwidth}{!}{
\begin{tabular}{lcccccc|cccccc}
\toprule
\textbf{Methods} & \multicolumn{6}{c|}{\textbf{Mapping}} & \multicolumn{6}{c}{\textbf{Functional Graph}} \\
\cmidrule(lr){2-7}\cmidrule(l){8-13}
& \multicolumn{2}{c}{Objects ($\uparrow$)} & \multicolumn{2}{c}{Inter.\ Elements ($\uparrow$)} & \multicolumn{2}{c|}{Overall Nodes ($\uparrow$)}
& \multicolumn{2}{c}{Node Assoc. ($\uparrow$)} & \multicolumn{2}{c}{Edge Pred. ($\uparrow$)} & \multicolumn{2}{c}{Overall Triplets ($\uparrow$)} \\
& R@3 & R@10 & R@3 & R@10 & R@3 & R@10
& R@5 & R@10 & R@5 & R@10 & R@5 & R@10 \\
\midrule
OpenFunGraph  & 81.1 & 87.8 & {71.0} & {79.5} & 73.0 & 82.8
              & {68.3} & {73.0} & \textbf{88.1} & \textbf{96.2} & \textbf{60.4} & {70.3} \\
\midrule
FunFact (Modified) & \textbf{90.5} & \textbf{95.2} & \textbf{76.4} & \textbf{88.7} & \textbf{81.1} & \textbf{90.9} & \textbf{76.4} & \textbf{83.1} & {75.8} & {91.4} & {57.9} & \textbf{75.9} \\
\bottomrule
\end{tabular}
}
\label{tab:fine_tune_on_scenefun3d}
\end{table*}
\begin{table*}
\centering
\caption{\textbf{Evaluating side effects of the pipeline adaptation.} To determine whether forcing knob and handle detection negatively affect overall mapping and functional prediction performance, we evaluate both the original and modified pipelines on \funthor.  The modified pipeline significantly improves recall for interactive elements (69.5\% $\to$ 87.9\%) and functional relations (49.3\% $\to$ 58.1\%), but substantially decreases precision (31.9\% $\to$ 16.0\%) and F1 score (38.7\% $\to$ 25.1\%), confirming that the modification negatively affects the overall performance. Nevertheless, the modified pipeline still outperforms OpenFunGraph in F1 score (25.1\% vs. 16.0\%).}
\renewcommand{\arraystretch}{1.1}
\resizebox{0.65\textwidth}{!}{
\begin{tabular}{lccc|ccc}
\toprule
\multirow{3}{*}{\textbf{Methods}} &  \multicolumn{3}{c}{\textbf{Mapping}}  & \multicolumn{3}{c}{\textbf{Functional Graph}}  \\
\cmidrule(lr){2-4} \cmidrule(lr){5-7}
& \multicolumn{3}{c|}{Recall @ 3 $(\uparrow)$} & \multicolumn{1}{c}{Prec. [\%]} & \multicolumn{1}{c}{Recall [\%]} & \multicolumn{1}{c}{F1 [\%]} \\
& \multicolumn{1}{c}{Objects} & \multicolumn{1}{c}{Inter.\ Elem.} & \multicolumn{1}{c|}{Overall Nodes} & $(\uparrow)$ & $(\uparrow)$ & $(\uparrow)$ \\
\midrule
OpenFunGraph        & 54.6 & 41.1 & 51.2 & 23.4 & 12.2 & 16.0 \\
\midrule
FunFact (Original)      & \textbf{68.2} & {69.5} & {68.5} & \textbf{31.9} & 49.3 & \textbf{38.7} \\
FunFact (Modified)  & \textbf{68.2} & \textbf{87.9} & \textbf{73.1} & 16.0 & \textbf{58.1} & 25.1 \\
\bottomrule
\end{tabular}
}
\label{tab:eval_fine_tuned}
\end{table*}

\section{VLM Prompt for Scene Analysis}
\label{app:prompt_scene_analysis}

We provide the complete prompt used for scene analysis in our mapping pipeline. This prompt instructs the vision-language model to identify functional objects and their interactive parts from RGB images of indoor scenes, including bounding box localization and part enumeration.
\begin{tcolorbox}[
  colback=white!0,
  colframe=gray!50,
  boxrule=0.5pt,
  left=2mm, right=2mm, top=2mm, bottom=2mm,
  breakable
]
\scriptsize\ttfamily
You are an expert at identifying functional objects in an RGB image of an indoor scene. Your task is to identify all functional objects and their interactive parts or interfaces that have potential functional relationships with other objects in the scene. A functional relationship means that the state change of one object will affect the state of other objects, or one object can be used to change the state of other objects. For example, turning on a light switch will turn on lights, turning a stove knob will turn on a burner, and using a key can open a specific lock or door.
\par\vspace{0.5em}
\#\# Extra Guidelines \\
1. If a handle appears rounded, resembles a knob, and is attached to a compartment to aid in opening or closing it (such as a cabinet or drawer), label this part as `knob handle`. \\
2. For all other handles that are functional components of an object (such as a door handle or drawer handle, except when covered in point 1), simply use `handle` as the part name, without specifying the handle type. \\
\par\vspace{0.5em}
\#\# Input Format \\
- A single RGB image of an indoor scene. \\
\par\vspace{0.5em}
\#\# Output Format \\
Your response should be a structured JSON object containing a list of identified objects. For each object, provide:
\par\vspace{0.5em}
\begin{itemize}[label=\texttt{-}]

    \item name: A common name for the object (\eg, 'stove', 'light switch'). Avoid using 'and' or 'or' in the name.

    \item description: A specific description that helps identify the object in the scene. This can include:
    \begin{itemize}[label=\texttt{-}, left=1.5em]
        \item Spatial information (\eg, 'a cabinet above a sink').
        \item Feature descriptions (\eg, 'light switch with dual switches').
        \item Or both.
    \end{itemize}

    \item functional\_parts: A list of functional parts of the object (\eg, for a 'stove': ['burner', 'knob']).
    \begin{itemize}[label=\texttt{-}, left=1.5em]
        \item If the object has no functional parts, leave this empty.
        \item List multiple parts separately.
        \item Avoid using 'and' or 'or' in the functional parts.
    \end{itemize}

    \item bounding\_box: The bounding box coordinates in the format [x\_min, y\_min, x\_max, y\_max].
    \begin{itemize}[label=\texttt{-}, left=1.5em]
        \item Coordinates should be normalized to the range [0, 1] relative to the image dimensions.
        \item Origin (0, 0) is at the top-left corner.
        \item x increases to the right, y increases downwards.
        \item The bounding box should be tight around the object, excluding the background.
    \end{itemize}

\end{itemize}
\end{tcolorbox}

\section{LLM Prompt for Local Functional Proposal}
\label{app:prompt_local_proposal}

We provide the complete prompt used for generating local functional proposals between objects and their parts. This prompt guides the language model to propose plausible part-object and part-part functional connections, assign prior confidence scores, and determine whether each connection type is inherently one-to-one.

\begin{tcolorbox}[
  colback=white!0,
  colframe=gray!50,
  boxrule=0.5pt,
  left=2mm, right=2mm, top=2mm, bottom=2mm,
  breakable
]
\scriptsize\ttfamily
You are an expert in understanding functional connections between objects and their parts in indoor scenes. A functional connection means that the state change of one object will affect the state of other objects, or one object can be used to change the state of other objects. For example, turning on a light switch will turn on the lights, turning a stove knob will turn on a burner, and using a key can open a specific lock or door. You are given a JSON string describing a parent object and its potential interactable parts. Your task is to: (1) propose plausible, semantically distinct functional connections between the parent object and its parts, as well as among the parts themselves; (2) provide a confidence score for each functional connection; and (3) indicate whether each connection is ``one-to-one'' or not. \\
\par\vspace{0.3em}
You are required to output a list of functional proposals in JSON format. A functional proposal is made of a functional connection in triplet format \texttt{<first\_item> <interaction> <second\_item>}, a confidence score, and an ``is\_one\_to\_one'' flag. You should only output semantically different functional proposals, and do not repeat the same type of interaction for different subjects/objects with the same name. For example, for a cabinet with multiple handles, only output a single functional proposal ``handle pull to open cabinet''.
\par\vspace{0.3em}

After proposing the function connection, you are required to give a confidence score of this connection semantically, like how correct it sounds based on common knowledge. If you are confident about your proposal based on common sense, like ``handle - pull to open - cabinet'', give a high score above 0.9. If you are unsure about your proposal and need more data to validate it, but it can still exist in reality, assign a low confidence score of around 0.5. \\
\par\vspace{0.3em}

If you believe the proposed functional connection is inherently ``one-to-one'' in general, for example, a light switch usually controls a single light, not multiple lights, set the flag ``is\_one\_to\_on''. On the other hand, a cabinet may contain multiple handles, and all of its handles can pull to open it. Therefore, the connection ``handle pull to open cabinet'' is not ``one-to-one''. \\
\par\vspace{0.3em}

The given JSON comes from a mapping pipeline, which may contain detection errors and misannotation. If the parent object contains an illogical part, for example, a faucet has a part named ``lock'', treat the part as a detection error and do not propose any functional connection for it. \\
\par\vspace{0.3em}

\#\# Pay attention
\begin{itemize}[label=\texttt{-}]
    \item Do not use passive voice in the ``interaction'' field. Use active voice instead. Swap the first and second items to avoid passive voice. For example, instead of ``light is turned on/off by switch'', use ``switch turns on/off light''.
\end{itemize}

\par\vspace{0.3em}

\#\# Input format

A JSON describing an object and its potential interactable parts. It has the following fields:
\begin{itemize}[label=\texttt{-}]
    \item object\_name: The name of the parent object.
    \item object\_description: A short description of the parent object.
    \item contained\_parts: List of part names that the parent object contains.
\end{itemize}

\par\vspace{0.3em}

\#\# Output format

A JSON of a list of your functional proposals. Each proposal should contain the following fields:
\begin{itemize}[label=\texttt{-}]
    \item first\_item\_name: Name of the first item of this proposal. It should be one of the part names from the given JSON input.
    \item second\_item\_name: Name of the second item of this proposal. It should be one of the part names or the parent object name from the given JSON input.
    \item interaction: A short sentence describing the action imposed by the first item on the second item (\eg, ``pull to open'', ``turns on/off'', ``controls'').
    \item confidence: Prior confidence score for this connection in general. The range is [0, 1].
    \item is\_one\_to\_one: Whether this connection type is inherently one-to-one.
\end{itemize}

\par\vspace{0.3em}

Your task is to propose plausible functional proposals between the object and its parts and among the parts themselves, give a confidence score of your functional proposal, and indicate whether such a connection is ``one-to-one'' or not.

\end{tcolorbox}

\section{LLM Prompt for Remote Functional Proposal}
\label{app:prompt_remote_proposal}

We provide the complete prompt used for generating object-object functional proposals. This prompt instructs the language model to identify inter-object functional connections, assign prior confidence scores, and determine the properties of each connection (one-to-one relationships and proximity requirements).

\begin{tcolorbox}[
  colback=white!0,
  colframe=gray!50,
  boxrule=0.5pt,
  left=2mm, right=2mm, top=2mm, bottom=2mm,
  breakable
]
\scriptsize\ttfamily
You are an expert at identifying functional connections between objects in indoor scenes. A functional connection means that the state change of one object affects the state of other objects, or one object can be used to change the state of other objects. For example, turning on a light switch turns on the lights; rotating a stove knob activates on a burner; and using a key can unlock a specific lock or door. You are given a JSON string describing a list of objects. Your task is to: (1) propose plausible, semantically distinct functional connections between these objects; (2) provide a confidence score for each functional connection; (3) indicate whether each connection is ``one-to-one'' or not; and (4) indicate whether each connection is a local and require close proximity between the two objects.

You are required to output a list of functional proposals in JSON format. A functional proposal is made of a functional connection in a triplet format \texttt{<first\_item> <interaction> <second\_item>}, a confidence score, and an ``is\_one\_to\_one'' flag. You should only output semantically different functional proposals, and do not repeat the same type of interaction for different subjects/objects with the same name. For example, if there are multiple switches and lights in the object list, only output a single functional proposal ``switch turn on/off light''.

After proposing the function connection, you are required to give a confidence score of this connection semantically, like how correct it sounds based on common knowledge. If you are confident about your proposal based on common sense, like ``power outlet - provide power to - TV'', give a high score above 0.9. If you are unsure about your proposal and need more data to validate it, but it can still exist in reality, assign a low confidence score of around 0.5.

If you believe the proposed functional connection is inherently ``one-to-one'' and exclusive in general, for example, a light switch usually controls a single light, not multiple lights, set the flag ``is\_one\_to\_one''. On the other hand, a power socket may provide power to multiple appliances; therefore, the connection ``power socket provides power to appliances'' is ``one-to-many'' and not ``one-to-one''.

If you believe that the proposed functional connection requires close proximity between the two objects to function properly, for example, a ``faucet pours water into sink'' requires the faucet to be close to the sink, set the flag ``is\_local'' to true. If the functional connection can work regardless of the distance between the two objects, for example, a ``remote control operates TV'' can work from a distance, set the flag ``is\_local'' to false.

\#\# Pay attention
\begin{itemize}[label=\texttt{-}]
    \item Do not use passive voice in the ``interaction'' field. Use active voice instead. Swap the first and second items to avoid passive voice. For example, instead of ``light is turned on/off by switch'', use ``switch turns on/off light''.
\end{itemize}

\#\# Input format

A JSON describing a list of objects:
\begin{itemize}[label=\texttt{-}]
    \item objects: A list of objects. Each object contains the following fields:
    \begin{itemize}[label=\texttt{-}, left=1.5em]
        \item name: Name of the object.
        \item description: A short description of the object.
    \end{itemize}
\end{itemize}

\#\# Output format

A JSON of a list of your functional proposals. Each proposal should contain the following fields:
\begin{itemize}[label=\texttt{-}]
    \item first\_item\_name: Name of the first item of this proposal. It should be one of the object names from the given JSON input.
    \item second\_item\_name: Name of the second item of this proposal. It should be one of the object names from the given JSON input.
    \item interaction: A short sentence describing the action imposed by the first item on the second item (\eg, ``pull to open'', ``turns on/off'', ``controls'', ``provide power to'', etc.).
    \item confidence: Prior confidence score for this connection in general. The range is [0, 1].
    \item is\_one\_to\_one: Whether this connection type is inherently one-to-one.
    \item is\_local: Whether this connection is local (\ie, the two objects need to be close to each other).
\end{itemize}

Your task is to propose plausible, semantically distinct functional proposals between the given objects, give confidence scores of your functional proposals, indicate whether such a connection is ``one-to-one'' or not, and indicate whether the connection is local or not.
\end{tcolorbox}

\end{document}